\PassOptionsToPackage{numbers,sort&compress}{natbib} 
\documentclass[preprint,a4paper]{elsarticle}
\usepackage{amssymb}
\usepackage{amsthm}
\usepackage{lineno}
\usepackage{amsmath,nccmath}
\usepackage{float}
\usepackage{subfig}
\usepackage{mathtools}
\usepackage{multirow}
\usepackage{tikz}
\usetikzlibrary{bayesnet}
\usetikzlibrary{arrows}
\usetikzlibrary{backgrounds}
\usepackage{url}
\usepackage{array}
\newcolumntype{P}[1]{>{\centering\arraybackslash}p{#1}}
\newcolumntype{M}[1]{>{\centering\arraybackslash}m{#1}}

\journal{Mechanical Systems and Signal Processing}

\begin{document}
	
	
\begin{frontmatter}
\title{On the hierarchical Bayesian modelling \\ of frequency response functions}
\author[add1]{T.A.\ Dardeno\corref{mycorrespondingauthor}}
\cortext[mycorrespondingauthor]{Corresponding author}
\ead{t.a.dardeno@sheffield.ac.uk}
\author[add1]{K.\ Worden}
\author[add1]{N.\ Dervilis}
\author[add1]{R.S.\ Mills}
\author[add2]{L.A.\ Bull}
\address[add1]{Dynamics Research Group, Department of Mechanical Engineering, \\ University of Sheffield, Sheffield S1 3JD, UK}
\address[add2]{Department of Engineering, University of Cambridge, CB3 0FA, UK}

\begin{abstract}
	Structural health monitoring (SHM) strategies seek to evaluate, predict, and maintain structural integrity, to improve the safety and design service life of structures in operation. Many of these strategies involve monitoring changes in structural dynamics, as damage can affect modal properties and present as changes in the characteristics of the resonance peaks of the frequency response function (FRF). While recent advances have improved the safety and reliability of structures, a number of challenges remain, impeding the practical implementation and generalisation of these systems. Like damage, benign variations, such as those caused by changes in temperature or other environmental fluctuations, can affect dynamic properties, making it difficult to distinguish between damage and normal operating conditions. In addition, newly-deployed structures can have insufficient data to describe the normal operating conditions (i.e., data \emph{scarcity}), which can impair the development of data-based prediction models. Another common challenge is data loss (i.e., data \emph{sparsity}), which may result from transmission issues, sensor failure, a sample-rate mismatch between sensors, and other causes. Missing data in the time domain will result in decreased resolution in the frequency domain, which can impair dynamic characterisation. 
	
	For situations that may benefit from information sharing among datasets, e.g., population-based SHM of similar structures, the hierarchical Bayesian approach provides a useful modelling structure. Hierarchical Bayesian models learn statistical distributions at the population (or parent) and the domain levels simultaneously, to bolster statistical strength among the parameters. As a result, variance is reduced among the parameter estimates, particularly when data are limited. In this paper, a combined probabilistic FRF model is developed for a small population of nominally-identical helicopter blades, using a hierarchical Bayesian structure, to support information transfer in the context of sparse data. The modelling approach is also demonstrated in a traditional SHM context, for a single helicopter blade exposed to varying temperatures, to show how the inclusion of physics-based knowledge can improve generalisation beyond the training data, in the context of scarce data. These models address critical challenges in SHM, by accommodating benign variations that present as differences in the underlying dynamics, while also considering (and utilising), the similarities among the domains.

\end{abstract}

\begin{keyword}
	Structural health monitoring (SHM); Population-based SHM (PBSHM); Hierarchical Bayes; Multilevel models; Uncertainty quantification; Repeatability
\end{keyword}

\end{frontmatter}


\section{Introduction} \label{intro}
	The current work is focussed on developing vibration-based structural health monitoring (SHM) strategies, that utilise \emph{data sharing}, domain expertise, and/or physics-based knowledge, to enable information transfer among similar domains. Specifically, this work has involved the development of probabilistic frequency response function (FRF) models, using data collected from healthy, full-scale composite helicopter blades, with consideration for the similarities and differences among the data, and the resulting effects on the characteristics of the resonance peaks.
	
	Benign variations confounding monitoring systems is one of the major challenges in SHM, as even among nominally-identical structures, variations caused by manufacturing differences, ageing parts, and changes in testing conditions can introduce uncertainty in the underlying dynamics. One such example of the effects of manufacturing differences on dynamic properties was presented in \cite{Cawley1985}, where Cawley \emph{et al.}\ performed vibration testing on filament-wound carbon fibre-reinforced plastic (CFRP) tubes. Of the 18 tested, six tubes were considered `normal' (i.e., having the same microstructure, with a $\pm 45^{\circ}$ fibre winding angle) \cite{Cawley1985}. They found that the first and second natural frequencies of the `normal' tubes varied by as much as 4\% \cite{Cawley1985}. When the remaining 12 tubes were also considered (which had intentional defects in their microstructure, such as slightly misaligned fibres, changes in volume fraction, etc.), the natural frequencies varied by as much as 18\% \cite{Cawley1985}. Likewise, changes in bolt tightness can similarly cause variations in natural frequency. Zhu \emph{et al.}\ \cite{Zhu2013bolts} performed modal testing on an aluminium three-bay space-frame structure with bolted joints, before and after manually loosening several bolts to hand tight. They found that natural frequencies changed by as much as 8\%, depending on the mode considered \cite{Zhu2013bolts}. 
	
	Global variations, such as changes in ambient temperature, can also affect modal properties. For example, increased temperature may reduce stiffness (depending on the specific material properties of the structure, and the duration/temperature of exposure), which may decrease natural frequency. Colakoglu \cite{Colakoglu2008} tested a polyethylene fibre composite beam from \mbox{-10}$^{\circ}$C to 60$^{\circ}$C, and found that the first natural frequency decreased by 12\% over the measured temperature range. In addition, for the same polyethylene beam, Colakoglu \cite{Colakoglu2008} found that modal damping increased with rising temperature. They concluded that the relationships between temperature and natural frequency, and between temperature and damping, appeared to be functional and monotonic for the polyethylene beam, over the temperature range considered \cite{Colakoglu2008}. Similar effects of temperature on stiffness/natural frequency and damping have been noted for a magnesium alloy \cite{Watanabe2004} and a composite honeycomb structure \cite{Bai2018}. Accounting for these benign fluctuations is important for the practical implementation and generalisation of SHM technologies, as features commonly used for damage identification may be sensitive to harmless changes as well as damage \cite{Worden2002NoveltyDI,HoonSohn,Alampalli,Cawley}. (Indeed, it is well known that structural damage can reduce stiffness, often manifesting as a reduction in natural frequency. Benign variations can then either mimic or mask damage, depending on whether they exhibit a stiffening or softening effect \cite{Alampalli,Cawley,HoonSohn}. Similar effects have been noted with damping, and multiple studies have shown that damping tends to increase with damage, particularly with crack growth \cite{Cao_2017}.) 
	
	Data \emph{scarcity} and \emph{sparsity} present additional challenges for SHM systems that rely on machine learning. Data scarcity refers to incomplete information regarding the damage- or normal-condition states of structures, particularly those newly in operation, and can impair model training and development. Likewise, sensing networks are prone to data loss (causing sparse data), because of sensor failure caused by harsh environmental conditions or insufficient maintenance. Transmission issues make wireless-sensing networks particularly susceptible to loss, and can be caused by large transmission distances between the sensors and base station \cite{PeiIMAC2005}, software/hardware problems \cite{Meyer2010}, and other issues such as weather changes, interference from nearby devices, or installation difficulties \cite{Bao2013}. Furthermore, modern systems that produce large amounts of high-resolution data can suffer losses resulting from a data transfer bottleneck \cite{Maleklo2022}. Significant losses higher than 30\% have been reported \cite{Meyer2010,Bao2013,Lynch2006}, and a 0.38\% data loss was found to have similar effects on power spectral density (PSD) as 5\% additive noise \cite{nagayama2007}. Differences in sample rate among sensors could have a similar presentation, with the data captured at a lower rate seemingly missing data relative to that captured at a higher rate. 
	
	Population-based SHM (PBSHM) addresses these challenges, by supporting knowledge transfer (e.g., damage or normal-condition states) within a population of structures, so that data-rich members can support those with less information. PBSHM strategies differ according to the similarity of the population members. \emph{Heterogeneous} populations \cite{gosliga2021foundations,gardner2021foundations} refer to those comprised of highly disparate members, such as different suspension bridge designs, and may require extensive processing such as domain adaptation prior to information sharing. The current work considers a \emph{homogeneous} population, comprised of nominally-identical members \cite{Bull_1}, where variabilities in their dynamics result from minor discrepancies in geometry, boundary conditions, and environmental/operational changes. Homogeneous populations can be represented using a general model, called a population \emph{form}, which attempts to capture the `essential nature' of the population and benign variations among the members \cite{Bull_1}. The form was first introduced in \cite{Bull_1}, where a conventional or single Gaussian process (GP) was applied to frequency response functions (FRFs), to develop a representation for a nominally-identical population of eight degree-of-freedom (DOF) systems. Then, to accommodate greater differences among the nominally-identical members, an overlapping mixture of Gaussian processes (OMGP) \cite{LAZAROGREDILLA20121386}, was used in \cite{Bull_1,Bull_3}, to infer multivalued wind-turbine power-curve data, with unsupervised categorisation of the data. The OMGP approach \cite{LAZAROGREDILLA20121386} was again used in \cite{Dardeno_1} to develop a population form for real and imaginary FRFs, obtained from four nominally-identical, full-scale, composite helicopter blades. Recently, hierarchical modelling was used to improve the predictive capability of simulated \cite{Bull_4} and in-service \cite{Bull_5} truck-fleet hazard models, and wind-turbine power curves \cite{Bull_5}. Specifically, the work presented in \cite{Bull_4, Bull_5} used \emph{partial pooling}, where data groups are treated as belonging to different domains, and each domain can be considered a realisation from global or population-level distributions. (This approach is in contrast to \emph{complete pooling}, where all data are considered as belonging to a single domain, or \emph{no pooling}, where each domain is treated as fully independent of the other domains). It was shown that when populations of structures were allowed to share correlated information, model uncertainty was reduced \cite{Bull_4, Bull_5}. In addition, domains with incomplete data were able to borrow statistical strength from data-rich groups \cite{Bull_4, Bull_5}. In \cite{LB_MRJ}, multilevel modelling with partial pooling was used to learn a 2D map of arrival times for waves propagating through a complex plate geometry, via GP regression, using a series of acoustic-emission experiments performed on the same plate but with differing experimental designs. Domain expertise regarding the positive gradient for the expected intertask functions was encoded into the model, via a linear mean function and appropriate priors \cite{LB_MRJ, Jones_1}. GP kernel hyperparameters were learnt at the domain level, to allow variation in the response surface among the different tests, and were correlated via a higher-level sampling distribution \cite{LB_MRJ}. 

	\subsection{Research aims of the current work} \label{intro_aims}
	The current work addresses critical SHM challenges, including uncertainties in the underlying dynamics of structures, and data-scarcity/sparsity issues, which can impair generalisation of SHM technologies. In line with the population forms developed in \cite{Bull_1, Bull_3, Dardeno_1, Bull_4, Bull_5}, this paper presents generalised, probabilistic FRF models that were developed for the helicopter blades in \cite{Dardeno_1}, using a hierarchical Bayesian structure. Two case studies are presented. The first case used FRFs collected from all four blades, at ambient laboratory temperature, with variations among the blades resulting from manufacturing differences (e.g., small discrepancies in material properties and geometry) and boundary conditions. Limited training data that did not fully characterise the resonance peaks were taken from two of the FRFs, while sufficient training data were taken from the remaining two FRFs, so that information could be shared with the data-poor domains via shared distributions over the parameters. This situation is representative of incomplete data in the time domain, which would reduce the number of spectral lines in the frequency domain. Independent models were generated for comparison, to visualise the variance reduction from the combined model. The second case addressed another challenge, typical in a traditional SHM context, where insufficient information regarding the normal operating condition of a structure can impede model generalisation. In this case, vibration data were collected from a single helicopter blade, at various temperatures, in an environmental chamber. A probabilistic FRF model was again developed using a hierarchical approach with partial pooling, except, for this model, each domain represented a different temperature state. Functional relationships between temperature and natural frequency, and between temperature and damping, were approximated via Taylor series expansion, and polynomial coefficients were learnt at the `population' level. A subset of temperature-varied FRFs were used to train the model. Higher-level distributions over the modal parameters, and the polynomial coefficients, were then used to generalise to temperature-varied FRFs not used to train the model. Model accuracy was evaluated by comparing the results to FRFs computed via measured vibration data.

	\subsection{Paper layout} \label{intro_layout}
	The layout of this paper is as follows. Section \ref{related_work} summarises existing research related to population-level monitoring of systems. Section \ref{novelty} outlines the novelty and contribution of the current work. Section \ref{theory} presents the theoretical basis for this research, including modal analysis and hierarchical Bayesian modelling techniques. Section \ref{dataset} briefly describes the datasets used to develop the models. Sections \ref{Case1} and \ref{Case2} discuss the models developed and analysis results for the first and second case studies, respectively. Conclusions are presented in Section \ref{conclusions}, and acknowledgements are presented in Section \ref{acknowledgements}.

\section{Related work} \label{related_work}
	
	Most literature related to population-level monitoring of engineering structures (or, in a more general SHM context, group-level monitoring of datasets), is focussed on \emph{transfer learning}, which aims to improve predictions in a \emph{target domain} given a \emph{source domain} with more complete information. Some cases have involved fine-tuning the classifiers and weights of a pre-trained convolutional neural network (CNN) according to a new dataset. For example, CNNs have been used to detect cracks, as in \cite{Wu2019,Dorafshan2018,Gao2018,Jang2019}, and other defects (e.g., corrosion, staining), as in \cite{Wu2019,Perez2019}. Other cases have used domain adaptation (DA), a subcategory of transfer learning, where the source and target domains have the same feature space, and the target domain is mapped onto a shared space. Predictions are then made from a single model. For example, domain-adversarial neural networks have been implemented for condition monitoring of a fleet of power plants \cite{Michau2019}, and for fault detection in a gearbox and three-story structure \cite{Ozdagli2020}. 
	
	Population-level modelling can also be viewed in the context of multitask learning (MTL), where multiple tasks are solved simultaneously, while considering similarities (and differences) across tasks. In MTL, a combined inference allows sharing of correlated information among domain-specific models, to improve accuracy in data-poor domains \cite{Bull_5,Sun2021}. In the context of modelling engineering infrastructure, MTL has been used as part of a multioutput Gaussian process (GP) regression to learn correlations between data obtained at adjacent sensors from the Canton Tower, to reconstruct missing information \cite{Wan2019}. (Note that using MTL in this context differs from the current work. In \cite{Wan2019}, missing time-domain data from temperature and acceleration sensors were reconstructed using MTL, and all data were sourced from a single structure. The current work considers frequency-domain data from multiple structures, as in the first case, and incorporates the relationships between temperature and domain-specific tasks into the model structure, as in the second case.) A similar approach was used in \cite{Li2021}, where GP regression assisted with missing data recovery from a faulty sensor, by capturing the correlation among the remaining sensors on a hydroelectric dam. Likewise, in \cite{Seshadri2022}, GPs were used to transfer information between spatial temperature or pressure profiles at axial stations within an aeroengine. 
	
	Hierarchical Bayesian modelling is an MTL framework, where (lower-level) domain-specific tasks are correlated by conditioning on (higher or population-level), \emph{shared} variables. An early introduction of these models to SHM was presented in \cite{Ballesteros2014}, where a hierarchical Bayesian framework was proposed for accommodating variability in the elastic modulus of simulated bar structures, and in \cite{Behmanesh2015}, which involved the model updating and uncertainty quantification of a numerical three-story shear-building model in the presence of environmental variations. Likewise, in \cite{Huang2015,Huang2019}, a multilevel hierarchical Bayesian model was developed to identify damage on a single structure by inferring stiffness loss from changes in modal parameters, given noisy, incomplete modal data. Modal parameters were learnt for different damage scenarios using a series of coupled linear regressions to approximate the eigenvalue problem \cite{Huang2015,Huang2019}. In \cite{Huang2019}, the similarities between acceleration responses at adjacent sensors were exploited for data-loss recovery, by modelling the measured signal as a linear combination of basis functions. Both cases presented in \cite{Huang2015,Huang2019} considered data from a single structure. (Note that the approaches presented in \cite{Ballesteros2014,Behmanesh2015,Huang2015,Huang2019} differ from the current work. The work presented in the current paper, in the first case study, models the entire (band-limited) FRF from \emph{multiple} structures, using a likelihood function based on modal parameters. The second case models the entire FRF using measurements at various temperatures from a single structure, and incorporates functional relationships between temperature and the modal parameters, enabling predictions at `unseen' temperature conditions.) 
	Recently, \cite{Poblete2022} used hierarchical Bayesian modelling to identify the electromechanical properties of energy harvesters, via individual and combined inference of experimentally-obtained FRFs. Population-level parameters were compared for partial- and complete-pooling approaches, successfully demonstrating that partial pooling results in improved estimates of population-level distributions compared to a classical Bayesian approach. The current work also utilises a hierarchical approach to probabilistic FRF modelling, but instead, the focus is on \emph{knowledge transfer} - whereby one or more members of a population or group have markedly fewer data. The current work also increases the complexity of the population (parameter) models, to move beyond exchangeable tasks, and predict the model parameters for previously-unobserved experiments. (In other words, the current work allows for the prediction of unseen models, as well as data.)
		
	Hierarchical Bayesian models were again used in \cite{DiFrancesco2021}, to estimate corrosion rates via data from multiple sensors on the same test structure, to support decision-making in the absence of complete data. In \cite{Papadimas2021} hierarchical Bayesian models were developed using strain measurements from multiple tensile tests to inform the material properties of the samples, in a manner that considered the inherent variability in material properties among the samples and uncertainties related to experimental repeatability. Likewise, in \cite{Dhada2020}, hierarchical Gaussian mixture models were used for combined inference in a simulated population of structures, for the purpose of damage identification. Another recent development was presented in \cite{Sedehi2023}, where a physics-informed GP model was developed using measured time-domain vibration data, with accommodation for temporal variability accomplished by partitioning the data into multiple subsegments, and applying a hierarchical Bayesian approach.

\section{Novelty and Contribution} \label{novelty}
	Unlike previous efforts, the current work is concerned with developing a probabilistic FRF model, using an FRF equation based on modal parameters (natural frequency, mode shape, and damping) as the likelihood function. As with \cite{Bull_4, Bull_5, LB_MRJ}, a hierarchical Bayesian (or MTL) framework was used. However, the focus is data sharing in the presence of low-resolution frequency information (i.e., given \emph{sparse} data). The inability to localise both time and frequency to a fine resolution, per the uncertainty principle, means that missing time-domain data (e.g., acceleration, velocity) measured from SHM sensors will result in fewer spectral lines in the frequency domain. This decreased frequency resolution could impair proper characterisation/identification of modal peaks, which may be features of interest in a damage-detection application. By using a combined-inference approach, structures whose FRFs have many spectral lines in a band of interest can lend statistical support to those whose FRFs are limited by missing data. In addition, the current work incorporated functional relationships to describe changes in environmental conditions. This inclusion of functional relationships allows for extrapolation to temperature states not used in model training, which increases the amount of normal-condition information available, to address data-scarcity challenges. In contrast to \cite{LB_MRJ}, parameters were learnt over the experimental campaign, rather than hyperparameters, giving greater physical interpretability of the results. In contrast to previous work, the current work focusses on information transfer for modal analysis and meta-analysis prediction for future models, rather than data pooling alone. Beyond conventional multilevel modelling and partial pooling, the intent is to learn how parameters vary over the population, even in the absence of complete data - an area that has not yet been addressed in modal analysis. Again, the current work allows for the prediction of new models, in addition to unseen data.

\section{Background theory} \label{theory}
	
	In this work, combined probabilistic FRF models were developed using a hierarchical Bayesian modelling approach, to support SHM research. This section provides an overview of the methodologies used, including hierarchical multilevel modelling from a Bayesian perspective, linear modal analysis, FRF estimation, and model evaluation.
	
	\subsection{Hierarchical Bayesian modelling} \label{theory_HBayes}

	Hierarchical models can be used to make combined inferences, whereby domains are treated as separate; but, at the same time, it is assumed that each domain is a realisation from a common latent model. This modelling structure involves \emph{partial pooling}, and is beneficial in that population-level distributions are informed by the full dataset, comprised of multiple domains. In partial pooling, certain parameters are permitted to vary between domains (i.e., \emph{varying parameters}); which are correlated by conditioning on parent variables at the population level. In the current work, the natural frequency would be a varying parameter. Other parameters can be considered \emph{shared} among members of a population (e.g., additive noise) and are learnt at the population level (these shared variables can still be sampled from parent distributions, which are also learnt at the population level). In contrast, a complete-pooling approach would consider all population data as having originated from a single source, while a no-pooling approach would involve fitting a single domain independently from the other domains. 

	Hierarchical models with partial pooling can be applied in a variety of SHM contexts, where a system can benefit from information sharing while preserving the uniqueness of the individual data sources. However, these models are particularly useful for PBSHM, and link intuitively to the concept of the population \emph{form}, which encompasses both the essential nature and variabilities among the members of a homogeneous population. Because parameters are allowed to vary at the domain level (as opposed to complete pooling), this approach can represent benign variations within a population. In addition, population-level variables are informed by the full dataset, rather than data from a single domain. This increase in statistical power is especially important in situations where one or more domains have limited data \cite{gelman2013bayesian,Bull_4,Bull_5}. In such cases, parameters from the data-poor domains exhibit shrinkage towards the population mean (therefore borrowing information from the other domains), which tightens the parameter variance \cite{gelman2013bayesian,Bull_4,Bull_5}. The differences among the data-pooling techniques are shown graphically in \mbox{Figure \ref{fig:pooling}}, using a simple linear regression example.
	
	\begin{center}
		{\includegraphics[width=\textwidth]{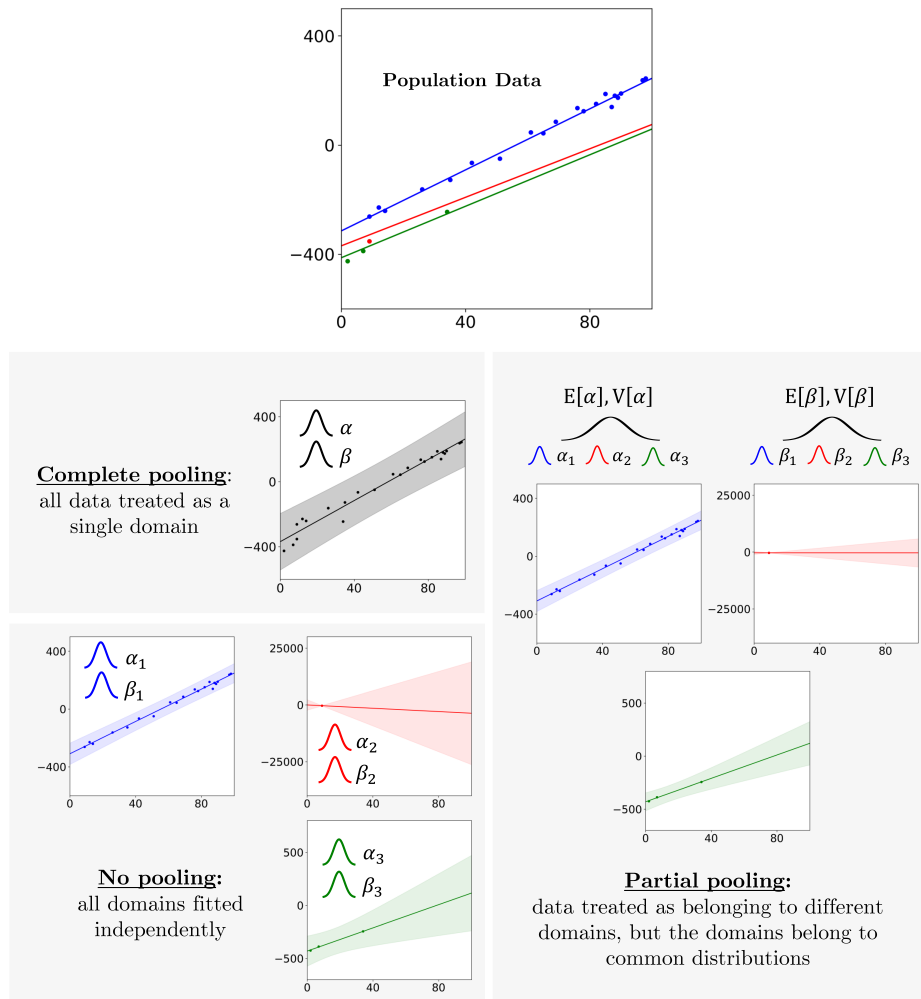}}
		\captionof{figure}{Comparison of data-pooling approaches, using a simple linear regression example.}
		\label{fig:pooling}
	\end{center}
	
	From a general perspective, data from a population comprised of $ K $ groups can be denoted via,
	
	\begin{equation}
		\left\{\mathbf{x}_{k},\mathbf{y}_{k}\right\}_{k=1}^K =
		\left\{\left\{x_{ik},y_{ik}\right\}_{i=1}^{N_k}\right\}_{k=1}^K
		\label{eq:PopulationData_1}
	\end{equation}
	
	\noindent where $ \mathbf{y}_{k} $ is the target response vector for inputs $ \mathbf{x}_{k} $, and $ \left\{x_{i,k},x_{i,k}\right\} $ are the $i$th pair of observations in group $ k $ \cite{Bull_5}. Each group is comprised of $ N_k $ observations, giving a total of $ \sum_{k=1}^{K}N_k $ observations \cite{Bull_5}. The objective is to learn a set of $ K $ predictors (one for each domain), related to the regression task, where the tasks satisfy,
	
	\begin{equation}
		\left\{y_{ik} = f_k\left(x_{ik}\right) + \epsilon_{ik}\right\}_{k=1}^{K}
		\label{eq:PopulationData_2}
	\end{equation}
		
	\noindent In other words, for each  $ i^{th} $ observation, the output is determined by evaluating one of $ K $ latent functions, $ f_k\left(x_{i,k}\right) $, plus additive noise, $ \epsilon_{i,k} $ \cite{Bull_5}. 
	
	While each of the $ k $ groups can be learned independently, a combined inference can be used to take advantage of the full $ \sum_{k=1}^{K}N_k $ population dataset. For example, consider a population that can be expressed using $ K $ linear regression models, as in Figure \ref{fig:pooling},
	
	\begin{equation}
		\left\{\mathbf{y}_{k} = \boldsymbol{\alpha}_k + \boldsymbol{\beta}_k\mathbf{x}_k + \boldsymbol{\epsilon}_k\right\}_{k=1}^{K}
		\label{eq:HierarchicalLinearRegression_1}
	\end{equation}
	
	\noindent where $ \boldsymbol{\alpha}_k $ and $ \boldsymbol{\beta}_k $ are the intercept and slope for domain $ k $, respectively; $ \mathbf{x}_k $ is a vector of inputs for the $k$th domain, with length $ N_k $; and $ \boldsymbol{\epsilon}_k $ is the noise vector for the $k$th domain, with length $ N_k $, and is assumed to be normally-distributed. Then, the likelihood of the target response vector is given as,
	
	\begin{equation}
		\mathbf{y}_{k} | \mathbf{x}_{k} \sim \mathcal{N}\left(\boldsymbol{\alpha}_k + \boldsymbol{\beta}_k\mathbf{x}_k,\sigma_k^2\right) 
		\label{eq:HierarchicalLinearRegression_2}
	\end{equation}

\begin{figure}[h]
		\centering
		\subfloat[\label{fig:LinearRegression_DGM1}]{
		\begin{tikzpicture}[latent/.style={circle, draw, minimum size=0.9cm}]
			\path (0,0) rectangle(3.5,3.5);
			\node[obs] (y) {$y_{i,k}$};
			\node[obs,right=1cm of y] (x) {$x_{i,k}$};
			\node[latent,above=0.75cm of y,xshift=1.2cm] (a) {$\boldsymbol{\alpha}_{k}$}; 
			\node[latent,above=0.75cm of y,xshift=-1.2cm] (b) {$\boldsymbol{\beta}_{k}$}; 
			\node[latent,above=0.75cm of a,xshift=-0.6cm] (mu_a) {$\mu_{\alpha}$}; 
			\node[latent,above=0.75cm of a,xshift=0.6cm] (sig_a) {$\sigma^2_{\alpha}$}; 
			\node[latent,above=0.75cm of b,xshift=-0.6cm] (mu_b) {$\mu_{\beta}$}; 
			\node[latent,above=0.75cm of b,xshift=0.6cm] (sig_b) {$\sigma^2_{\beta}$}; 
			\node[latent,left=1cm of y] (sig_y) {$\sigma_k^{2}$}; 
			\plate [inner sep=.25cm,yshift=.1cm,xshift=-.1cm]{plateN} {(y)(x)} {$i \in 1:N_k$};
			\edge {a,b,sig_y,x} {y} 
			\edge {mu_a,sig_a} {a} 
			\edge {mu_b,sig_b} {b} 
		\end{tikzpicture}
		}
		\subfloat[\label{fig:LinearRegression_DGM2}]{
		\begin{tikzpicture}[latent/.style={circle, draw, minimum size=0.9cm}]
			\path (0,0) rectangle(3.5,3.5);
			\node[obs] (y) {$y_{i,k}$};
			\node[obs,right=1cm of y] (x) {$x_{i,k}$};
			\node[latent,above=0.75cm of y,xshift=1.2cm] (a) {$\boldsymbol{\alpha}_{k}$}; 
			\node[latent,above=0.75cm of y,xshift=-1.2cm] (b) {$\boldsymbol{\beta}_{k}$}; 
			\node[latent,above=0.75cm of a,xshift=-0.6cm] (mu_a) {$\mu_{\alpha}$}; 
			\node[latent,above=0.75cm of a,xshift=0.6cm] (sig_a) {$\sigma^2_{\alpha}$}; 
			\node[latent,above=0.75cm of b,xshift=-0.6cm] (mu_b) {$\mu_{\beta}$}; 
			\node[latent,above=0.75cm of b,xshift=0.6cm] (sig_b) {$\sigma^2_{\beta}$}; 
			\node[latent,left=1.75cm of y] (sig_y) {$\sigma^{2}$}; 
			\plate [inner sep=.25cm,yshift=.1cm,xshift=-.1cm]{plateN} {(y)(x)} {$i \in 1:N_k$};
			\plate [inner sep=.25cm,yshift=.1cm,xshift=-.1cm]{plateK} {(plateN)(a)(b)} {$k \in 1:K$};
			\edge {a,b,sig_y,x} {y} 
			\edge {mu_a,sig_a} {a} 
			\edge {mu_b,sig_b} {b} 
	\end{tikzpicture}
	}
		\caption{DGM of linear regression model, (a) independent (no pooling) and (b) partial pooling.}
	\end{figure}
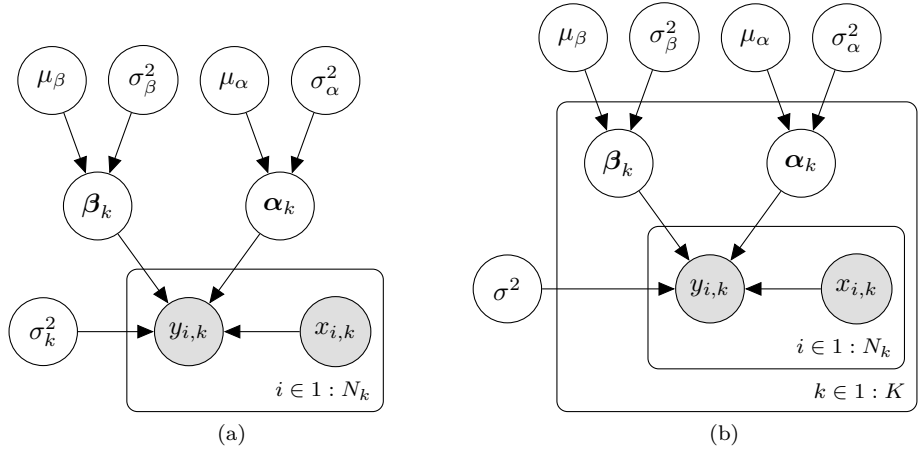

	\noindent A shared hierarchy of prior distributions can be placed over the slopes and intercepts for the groups $ k \in \left\{1, ... \,, K\right\} $, in line with a Bayesian framework. To allow information to flow between groups, parent nodes $ \left\{\boldsymbol{\mu}_\alpha,\boldsymbol{\sigma}^2_\alpha\right\} $ and $ \left\{\boldsymbol{\mu}_\beta,\boldsymbol{\sigma}^2_\beta\right\} $ can be learned at the population level. Note that sometimes, it may be appropriate to learn certain parameters at the population level rather than the domain level. If, for example, the same hardware was used for data collection within the population, one could assume a global noise variance ($ \sigma^2 $) that is the same for each domain. Consider the directed graphical models (DGMs) shown in Figures \ref{fig:LinearRegression_DGM1} and \ref{fig:LinearRegression_DGM2}. Figure \ref{fig:LinearRegression_DGM1} shows the DGM for the linear regression example, given an independent (no-pooling) approach. Each model is learnt independently and no information is shared. Slopes, intercepts, and noise variance are learnt individually for each $ k $ domain, but there is no plate surrounding them to denote repetition over the $ K $ domains. While the slopes and intercepts are sampled from shared parent nodes $ \left\{\boldsymbol{\mu}_\alpha,\boldsymbol{\sigma}^2_\alpha\right\} $ and $ \left\{\boldsymbol{\mu}_\beta,\boldsymbol{\sigma}^2_\beta\right\} $, these nodes are informed only by the data from a \emph{single} domain and any priors. In contrast, Figure \ref{fig:LinearRegression_DGM2} shows the DGM given a partial-pooling approach. The slopes and intercepts are indexed by $ k $, and plate notation is used to show that these nodes are repeated. Once again, shared parent nodes are outside of the plates and not indexed by $ k $, meaning that these are population-level variables, and information is permitted to flow between these and the domain-specific parameters. In other words, the parent nodes are now informed by the data from \emph{multiple} domains and prior information. The noise variance ($ \sigma^2 $) is shared among the domains, at the population level, as indicated by the absence of a $ k $ subscript, and its exclusion from the $ K $ plate.

	The current work considers a hierarchical probabilistic FRF model, that uses an accelerance FRF estimate (based on modal parameters), as the mean of the likelihood function. A brief introduction to modal analysis and FRF estimation is provided.
		
	\subsection{Modal analysis and FRF estimation} \label{theory_modal}
	The equation of motion for a multiple DOF system can be written as,
	
	\begin{equation}
		\mathbf{M}\ddot{\mathbf{u}}(t) + \mathbf{C}\dot{\mathbf{u}}(t) + \mathbf{K}{\mathbf{u}}(t) = \mathbf{z}(t)
		\label{eq:EOM}
	\end{equation}
	
	\noindent where $ \ddot{\mathbf{u}}(t) $, $ \dot{\mathbf{u}}(t) $, $ {\mathbf{u}}(t) $, and $ \mathbf{z}(t) $ are acceleration, velocity, displacement, and force, respectively. In most cases, the mass $ \mathbf{M} $, damping $ \mathbf{C} $, and stiffness $ \mathbf{K} $ matrices are coupled. For linear systems, and in the absence of viscous damping, the equation of motion can be decoupled, such that the system is represented by multiple single degree-of-freedom (SDOF) oscillators. This decoupling is performed via the eigenvalue expression,
	
	\begin{equation}
		[\mathbf{K}-\omega_n^2\mathbf{M}]\boldsymbol{\Psi} = 0
		\label{eq:eig}
	\end{equation}
	
	\noindent and yields the natural frequencies in radians, $ \omega_n $, and mode shapes, $ \boldsymbol{\Psi} $, of the system. 
	
	The physical equation of motion can then be cast in modal space to give the uncoupled modal equations, with modal coordinates $ {\mathbf{p}}(t) $, written as,
	
	\begin{equation}
		\boldsymbol{\Psi}^\text{T}\mathbf{M}\boldsymbol{\Psi}\ddot{\mathbf{p}}(t)+ \boldsymbol{\Psi}^\text{T}\mathbf{C}\boldsymbol{\Psi}\dot{\mathbf{p}}(t) + \boldsymbol{\Psi}^\text{T}\mathbf{K}\boldsymbol{\Psi}{\mathbf{p}}(t) = \boldsymbol{\Psi}^\text{T}\mathbf{z}(t)
		\label{eq:modalEOM}
	\end{equation}
	
	\noindent where,
	
	\begin{equation}
		\mathbf{u}(t) = \boldsymbol{\Psi}{\mathbf{p}}(t)
		\label{eq:phystomodal}
	\end{equation}
	
	FRFs aid visualisation of the natural frequency components of a system, and are computed by normalising the response signal at a given location to the excitation force. This work used the $ \text{H}_1 $ estimator to compute FRFs. For a response at location $ h $ resulting from excitation at $ j $, the $ \text{H}_1 $ estimator is computed as,
	
	\begin{equation}
		\mathbf{H}_{hj}(\omega) = \frac{\mathbf{G}_{zu}(\omega)}{\mathbf{G}_{zz}(\omega)}
		\label{eq:FRF}
	\end{equation}
	
	\noindent where,
	
	\begin{equation}
		\begin{split}
			& \mathbf{G}_{zu}(\omega) \overset{\Delta}{=} {\it \mathbb{E}} [\mathbf{S}_{u_h}(\omega) \mathbf{S}_{z_j}^*(\omega)] \\
			& \mathbf{G}_{zz}(\omega) \overset{\Delta}{=} {\it \mathbb{E}} [\mathbf{S}_{z_j}(\omega) \mathbf{S}_{z_j}^*(\omega)]
		\end{split}
		\label{eq:spectra}
	\end{equation}
	
	\noindent and,
	
	\begin{equation}
		\begin{split}
			& \mathbf{S}_{z_j}(\omega) \overset{\Delta}{=} \mathcal{F}[{\mathbf{z}_j}(t)] \\
			& \mathbf{S}_{u_h}(\omega) \overset{\Delta}{=} \mathcal{F}[{\mathbf{u}_h}(t)]
		\end{split}
		\label{eq:FFT}
	\end{equation}
	
	\noindent The asterisk $ * $ denotes complex conjugation, $ \omega $ is frequency in radians, and $ \mathcal{F} $ is a discrete Fourier transform (this work used a fast Fourier transform or FFT). The input force in the time domain at location $ j $ is $ {\mathbf{z}_j}(t) $, and the output response (i.e., acceleration, velocity, or displacement) in the time domain at $ h $ is $ {\mathbf{u}_h}(t) $.
	
	With assumed linear behaviour and proportional damping, the (complex) accelerance FRF (i.e., given acceleration response data) can also be estimated using modal parameters,
	
	\begin{equation}
		\mathbf{H}_{hj}(\boldsymbol{\omega}) = -\boldsymbol{\omega}^2 \sum_{m=1}^{M} \frac{A_{hj}^{(m)}}{\omega_{nat,m}^2-\boldsymbol{\omega}^2+2i\zeta_m\boldsymbol{\omega}\omega_{nat,m}}
		\label{eq:modalFRF}
	\end{equation}
	
	\noindent where $ A_{hj}^{(m)} $ is the residue for mode $ m $, defined as the product of the mass-normalised mode shapes at locations $ h $ and $ j $ ($ A_{hj}^{(m)} = \psi_{hm}\psi_{jm} $) \cite{wordennonlinearity}. The natural frequency associated with mode $ m $ is $ \omega_{nat,m} $, and the modal damping associated with mode $ m $ is $ \zeta_m $ \cite{wordennonlinearity}. The real and imaginary parts of the FRF can be computed independently, by multiplying Eq.\ (\ref{eq:modalFRF}) by its complex conjugate. This work considers FRFs from multiple domains (different structures in Case 1, and a single structure in Case 2 with data obtained at varying temperatures), with each domain indexed by $ k $. In each case, only one FRF is assigned to each domain, from a given measurement location (so, subscripts $ h $ and $ j $ can be neglected). Thus, the real and imaginary components of the FRF from the $k$th domain, given a vector of frequency inputs  $ \boldsymbol{\omega}_{k} $, can be estimated via,
	
	\begin{equation}
		\textrm{real}\left[\mathbf{H}_k\left(\boldsymbol{\omega}_{k}\right)\right] = -\boldsymbol{\omega}^2_{k} \sum_{m=1}^{M} \frac{A_{m}\left(\omega_{nat}^{2^{(k,m)}} - \boldsymbol{\omega}^2_{k}\right)}{\omega_{nat}^{4^{(k,m)}}+\boldsymbol{\omega}^4_{k} + 2\boldsymbol{\omega}^2_{k}\omega_{nat}^{2^{(k,m)}}\left(2\zeta_{(k,m)}^2-1\right)}
		\label{eq:modalFRFreal}
	\end{equation}
	
	\begin{equation}
		\textrm{imag}\left[\mathbf{H}_k\left(\boldsymbol{\omega}_{k}\right)\right] = -\boldsymbol{\omega}^2_{k} \sum_{m=1}^{M} \frac{-2A_{m}i\zeta_{(k,m)}\boldsymbol{\omega}_{k}\omega_{nat}^{{(k,m)}}}{\omega_{nat}^{4^{(k,m)}}+\boldsymbol{\omega}^4_{k} + 2\boldsymbol{\omega}^2_{k}\omega_{nat}^{2^{(k,m)}}\left(2\zeta_{(k,m)}^2-1\right)}
		\label{eq:modalFRFimag}
	\end{equation}
	
	Note that residues, $ A_{m} $, are not indexed by $k$ in Eq.\ (\ref{eq:modalFRFreal}). For the models presented in the case studies below, mode shapes were shared among the domains, to address identifiability concerns during sampling, and because mode shapes are often less sensitive to global variations compared to other modal parameters.
	
	The current work has focussed on the real part of the FRF, and Eq.\ (\ref{eq:modalFRFreal}) provided the mean of the likelihood function for the models developed. The residues, modal damping, and natural frequencies for each domain were learnt as parameters for the hierarchical model, with distributions over these parameters learnt as hyperparameters. Learning the real part of the FRF was sufficient for demonstrating the proposed technology. However, the imaginary part could be learnt using the same methods, with Eq.\ (\ref{eq:modalFRFimag}) providing the likelihood function, or could be inferred (at least in part), by exploiting the causal relationship between the real and imaginary components of the FRF \cite{wordennonlinearity}. 
	
	\subsection{Model evaluation for generalisation beyond training data (Case 2)} \label{theory_model_eval}
	The normalised mean-squared error (NSME) was calculated to evaluate the accuracy of the extrapolation to temperatures beyond the training data, for the second case presented herein, via

	\begin{equation}
		\text{NMSE} = \frac{100}{M\sigma^2_{\mathbf{y}}}\sum_{i=1}^{M}\left(y_i - y_i^*\right)
		\label{eq:NMSE}
	\end{equation}

	\noindent where $ M $ is the number of test data points, $ y $ is the test data, $ \sigma^2_{\mathbf{y}} $ is the variance of the test data, and $ y^* $ is the predicted function. Normalising to the test data variance allows for comparison of results on a consistent scale, regardless of signal magnitude. According to convention, an NMSE less than 5\% suggests that the model fits the data well.

\vspace{12pt} 
\section{Dataset summary} \label{dataset}
	The population dataset was comprised of vibration data collected from four healthy, nominally-identical, full-scale composite\footnote{Although the exact internal geometry of the blades is unknown, specifications for Gazelle helicopter blades have indicated that they are comprised of steel, fibreglass, and a honeycomb core \cite{Garinis2012}.} blades from a Gazelle helicopter (referenced in this paper as Blades 1-4). Data were collected using Siemens PLM LMS SCADAS hardware and software at the Laboratory for Verification and Validation\footnote{https://lvv.ac.uk/} (LVV) in Sheffield, UK. The first case used FRFs calculated from data that were collected on all four blades at ambient laboratory temperature, as described in \cite{Dardeno_1}. The second case used FRFs calculated from data that were collected on a single blade (Blade 1) at multiple temperatures in an environmental chamber.
	
	\subsection{Data collection at ambient laboratory temperature} \label{data_Case1}
	The first case used data collected at ambient laboratory temperature, with the blades in a fixed-free boundary condition, which was approximated by placing the root end of each blade in a substantiated strong-wall mount. Ten uniaxial 100 mV/g accelerometers were placed along the length of the underside of each blade. Note that the same accelerometers were used on each blade, and care was taken to ensure that they were attached to approximately the same locations on each blade. An electrodynamic shaker with force gauge was mounted to a fixture bolted to the laboratory floor and attached to the blade 0.575 metres from the root. The shaker was attached to the underside of the blade in the flapwise direction. A continuous random excitation was generated in LMS (note: LMS refers to Siemens PLM LMS SCADAS hardware and software) and applied to excite the blade up to 400 Hz, with a step size of 4.88e-02 Hz. Throughput time data were collected for each test, and the data were divided into 20 blocks. Hanning windows were applied and FRFs were computed for each time block, which were then averaged in the frequency domain. The experimental setup is shown in Figure \ref{fig:blade_ambient}, and the accelerometer positions on the blades are shown in Figure \ref{fig:sensor_layout}. 
	
\begin{figure}[h!]
		\centering
		\subfloat[\label{fig:FRFs_allblades_real}]{\includegraphics[width=0.8\columnwidth]{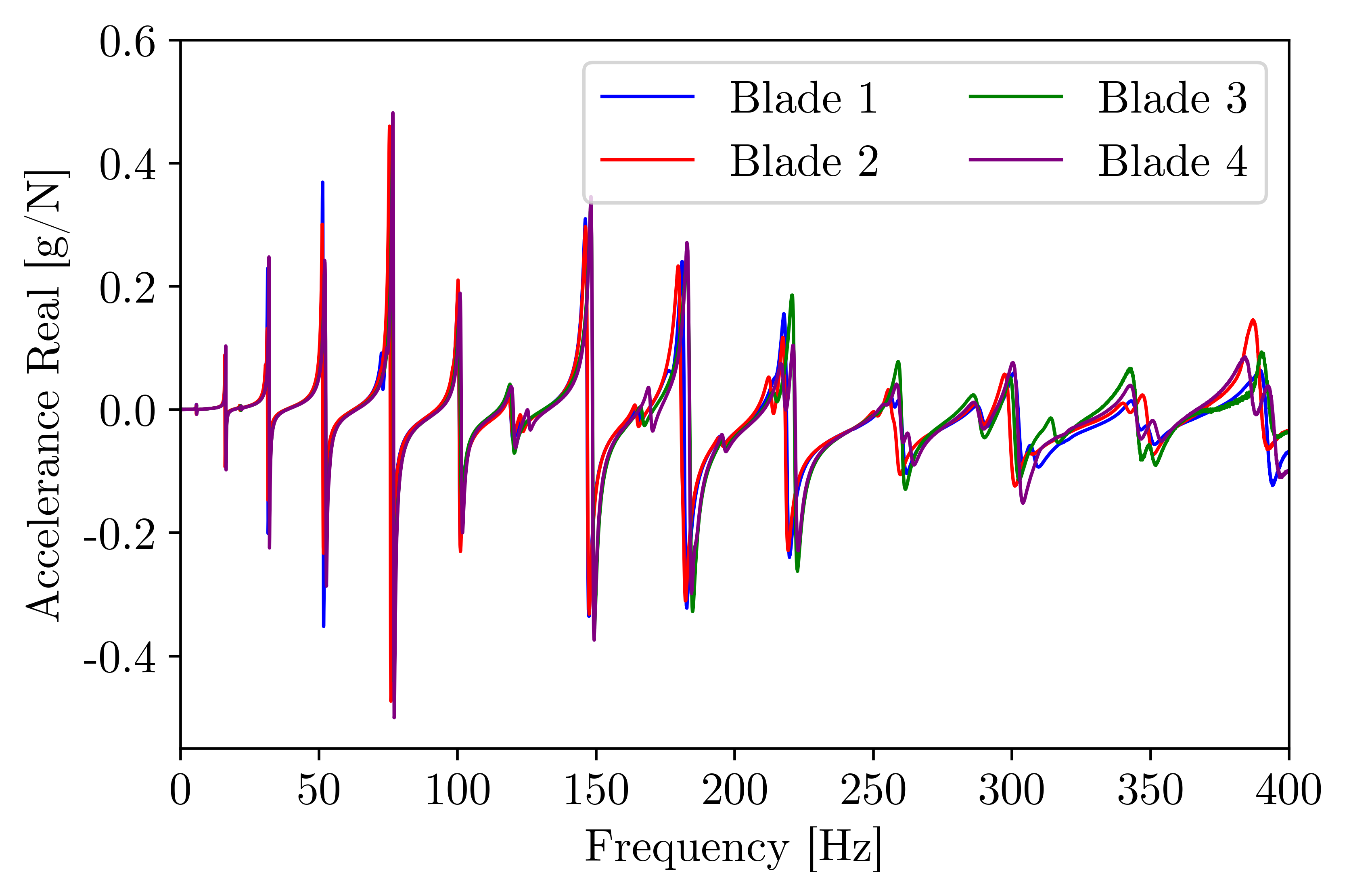}} \\
		\subfloat[\label{fig:FRFs_allblades_real_2DOF}]{\includegraphics[width=0.8\columnwidth]{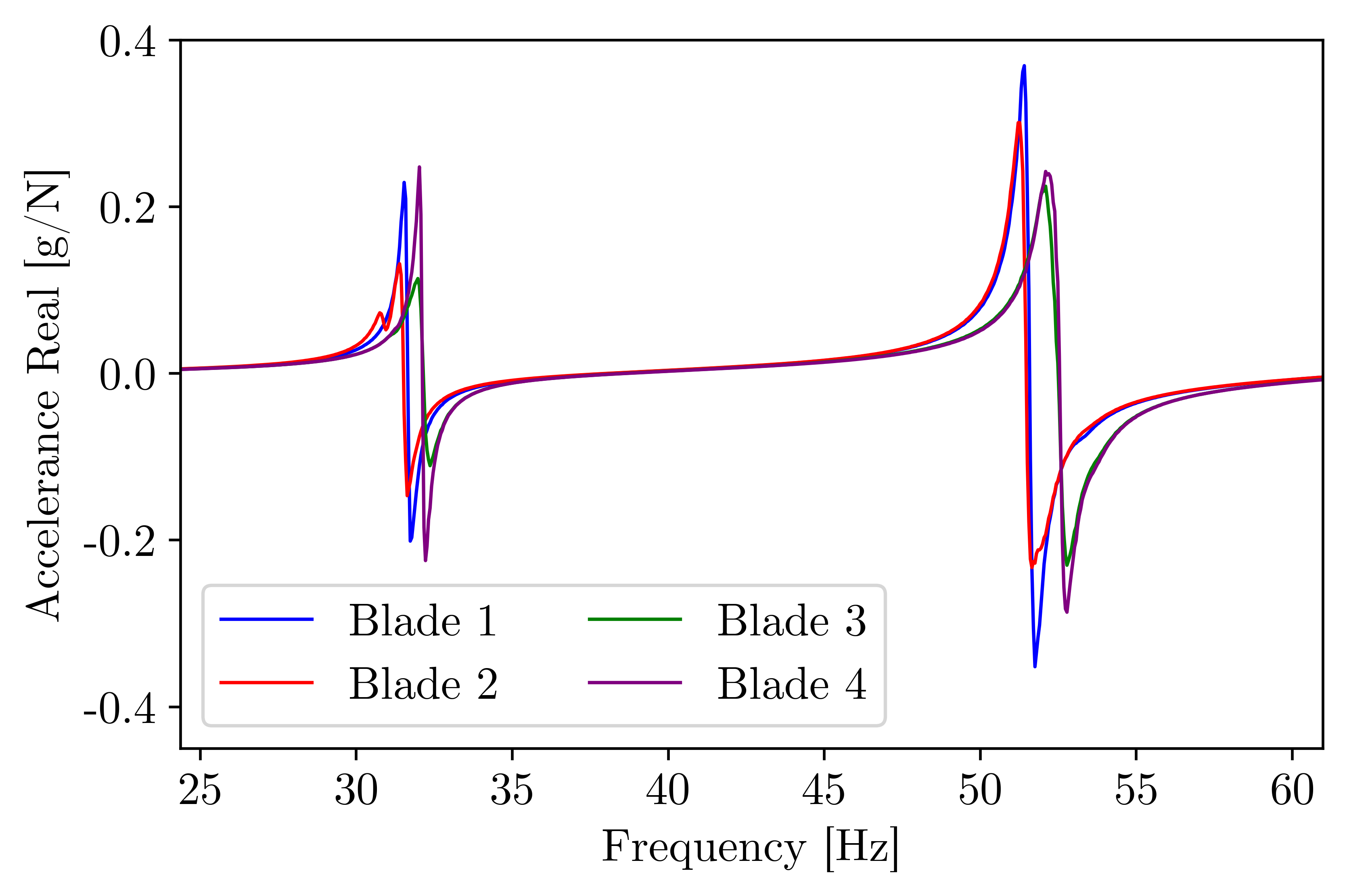}}
		\caption{Real components of the helicopter blades FRFs, (a) full bandwidth and (b) bandwidth of interest, at the drive-point location, from vibration data collected at ambient laboratory temperature.}
	\end{figure}

	To demonstrate the proposed technique and reduce the number of model hyperparameters (thus minimising model complexity and reducing computation time), a narrow frequency band was selected around two lightly-damped, well-separated modes. Specifically, the chosen bandwidth was between 24 and 61 Hz, with the fourth and fifth bending modes of the blades dominating the response in this band. Although other modes appear to have a small influence in this band, a 2DOF assumption was imposed. (This assumption results in smoothing of the FRF over the band, and might result in some loss of interpretability, but is acceptable for these preliminary analyses). The real part was modelled as a probabilistic FRF, using the FRF estimate from Eq.\ (\ref{eq:modalFRFreal}) as the mean of the likelihood function, as described in Case 1, presented in Section 6 of this paper. The real parts of the averaged FRFs for each blade, at the second accelerometer from the blade root (corresponding to the drive-point location), are shown in Figures \ref{fig:FRFs_allblades_real} and \ref{fig:FRFs_allblades_real_2DOF}. Figure \ref{fig:FRFs_allblades_real} shows the full measured bandwidth, and Figure \ref{fig:FRFs_allblades_real_2DOF} shows the FRF in the bandwidth of interest, between 24 and 61 Hz. 
	
	\begin{center}
		{\includegraphics[width=\textwidth, trim = {0 6cm 2cm 8cm},clip]{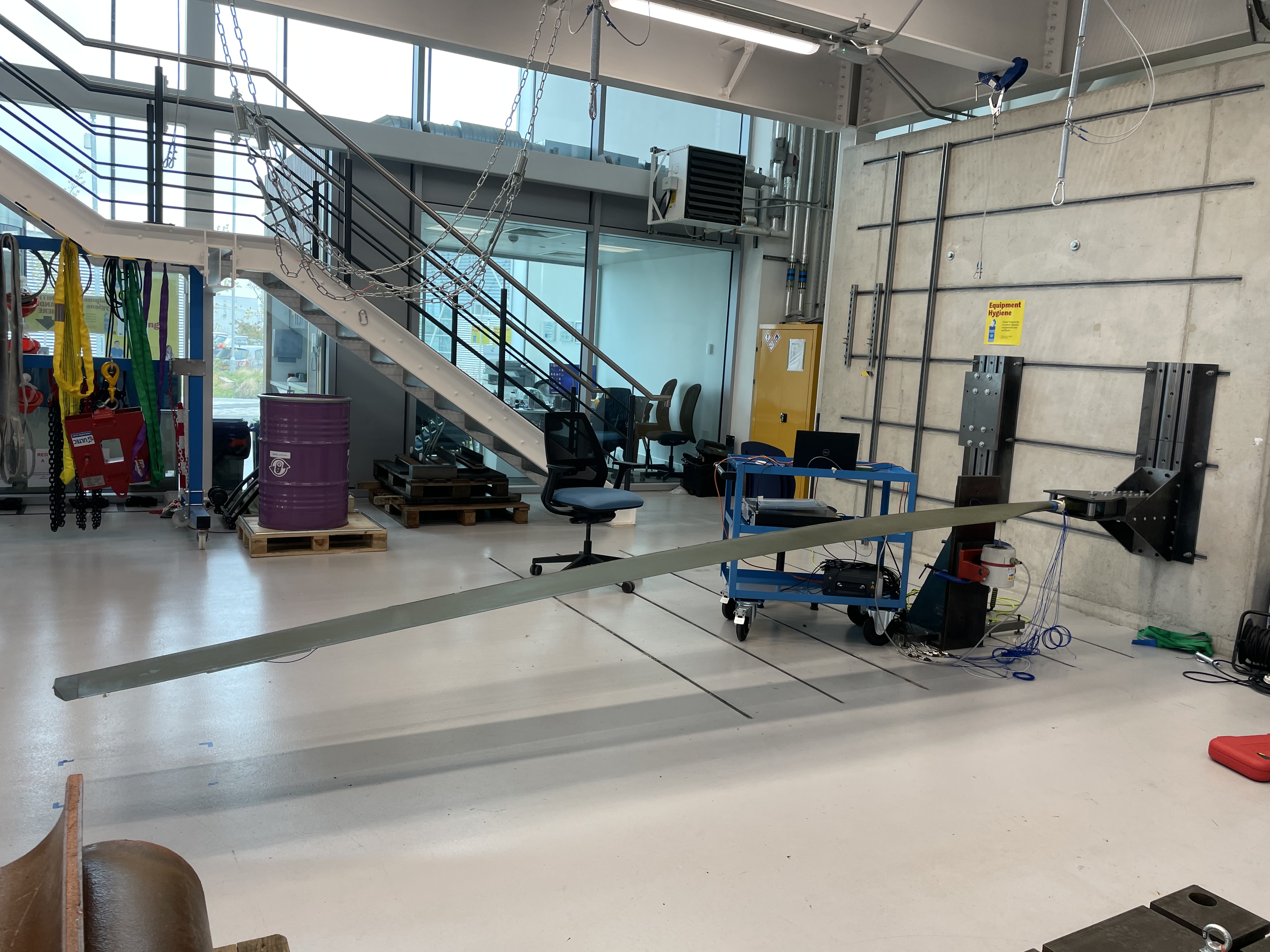}}
		\captionof{figure}{Helicopter blade in a substantiated wall mount.}
		\label{fig:blade_ambient}
	\end{center}

	\begin{center}
		{\includegraphics[width=\textwidth, trim = {2.8cm 0cm 2.5cm 0cm},clip]{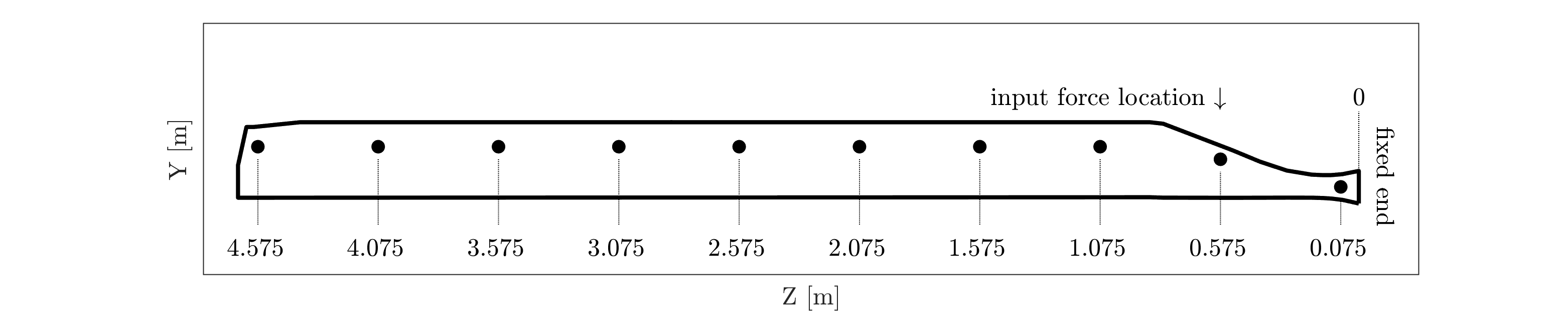}}
		\captionof{figure}{Sensor locations on the helicopter blades.}
		\label{fig:sensor_layout}
	\end{center}
	
	Figures \ref{fig:FRFs_allblades_real} and \ref{fig:FRFs_allblades_real_2DOF} show increasing variability with respect to frequency, which is an expected result, as higher-frequency modes are more sensitive to small physical changes than lower-frequency modes. For modes less than 80 Hz, the maximum frequency difference among the blades was approximately 2.5 Hz; for modes greater than 80 Hz, the maximum frequency difference was approximately 6.3 Hz. Note the grouping visible at several of the peaks, where Blades 1 and 2 appear closely aligned in frequency while Blades 3 and 4 appear closely aligned. These results are quite relevant for PBSHM. All of the helicopter blades are healthy, and represent a normal-condition state of the population. Consider a situation where only FRFs from one of the groupings are available for training a model (or FRFs from the other groups are missing data). The normal condition could be heavily biased towards the training set, and incoming FRFs could be flagged as damaged, even if they are healthy. Further details regarding the data collection and processing for these tests can be found in \cite{Dardeno_1}.

	\subsection{Data collection at various temperatures in an environmental chamber} \label{data_Case2}
\begin{figure}[!h]
		\centering
		{\includegraphics[width=1\textwidth, trim = {3cm 7.2cm 5cm 11.2cm},clip]{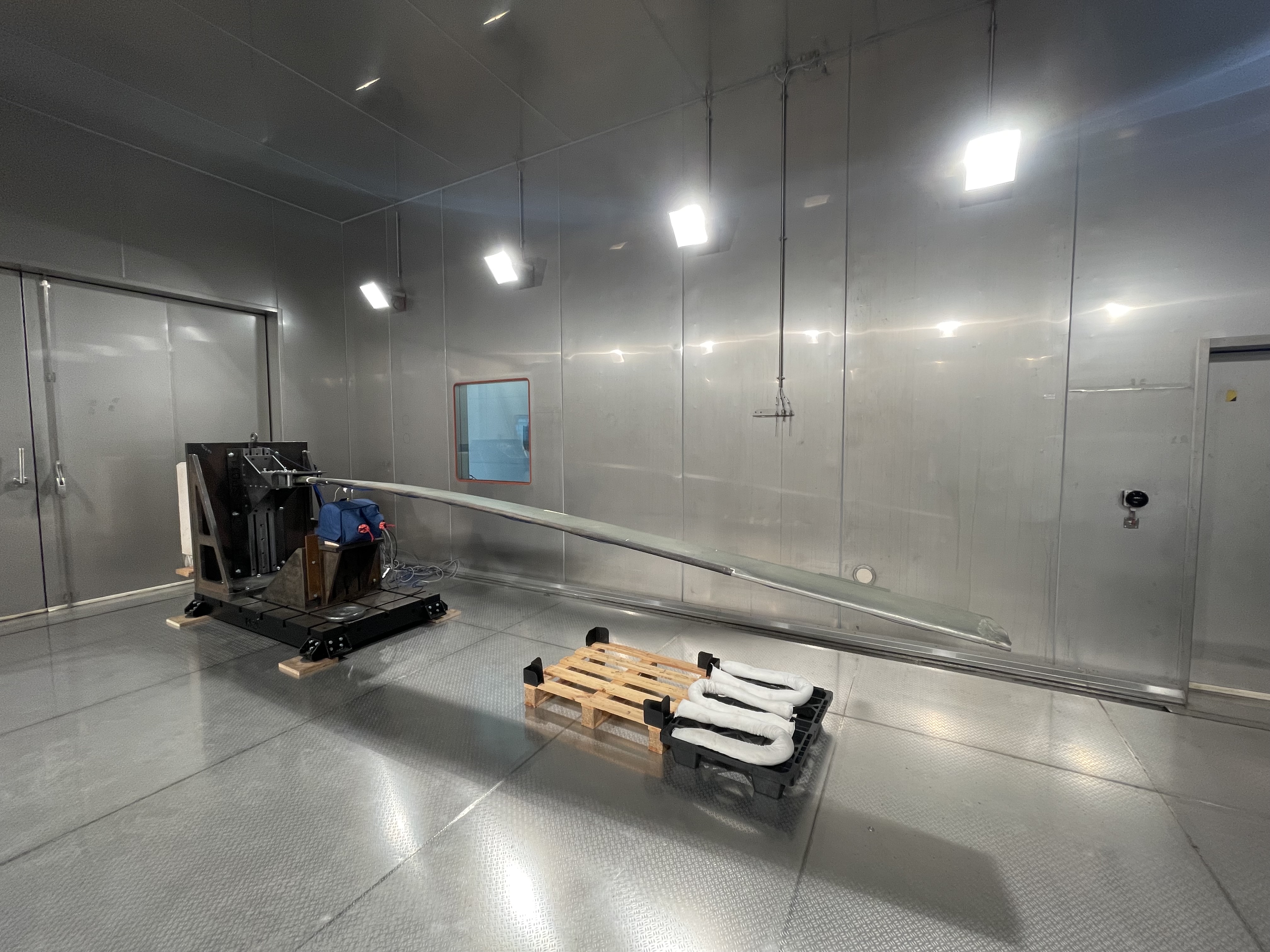}} \\
		\caption{Helicopter blade in an environmental chamber.}
		\label{fig:blade_temps}
	\end{figure}

	The second case used data from Blade 1, collected at temperatures ranging from -20 to 30$^{\circ}$C in increments of 5$^{\circ}$C in an environmental chamber. The blade was tested in an approximately fixed-free boundary condition, with the blade root mounted on a fixture that was substantiated with a large concrete block. These tests used the same accelerometers, sensor layout, data-acquisition parameters, and processing methods as the previous tests at ambient laboratory temperature. The shaker was mounted to the bottom plate of the test fixture, and the shaker and force gauge were connected to the underside of the blades to excite the blades in the flapwise direction. A thermal jacket was used to protect the shaker when testing at lower temperatures. Prior to data collection, the environmental chamber was set to each of the temperatures of interest, after which the blades were allowed to soak for at least two hours to reach the desired temperature. Throughput force and acceleration data were collected for each test. The data were then divided into 20 blocks, and a Hanning window was applied to each data block. FRFs were then computed for each block, and averaged in the frequency domain. The experimental setup, with the helicopter blade inside the environmental chamber, is shown in Figure \ref{fig:blade_temps}. 
	
\begin{figure}[h!]
		\centering
		\subfloat[\label{fig:FRFs_temps_real}]{\includegraphics[width=0.8\columnwidth]{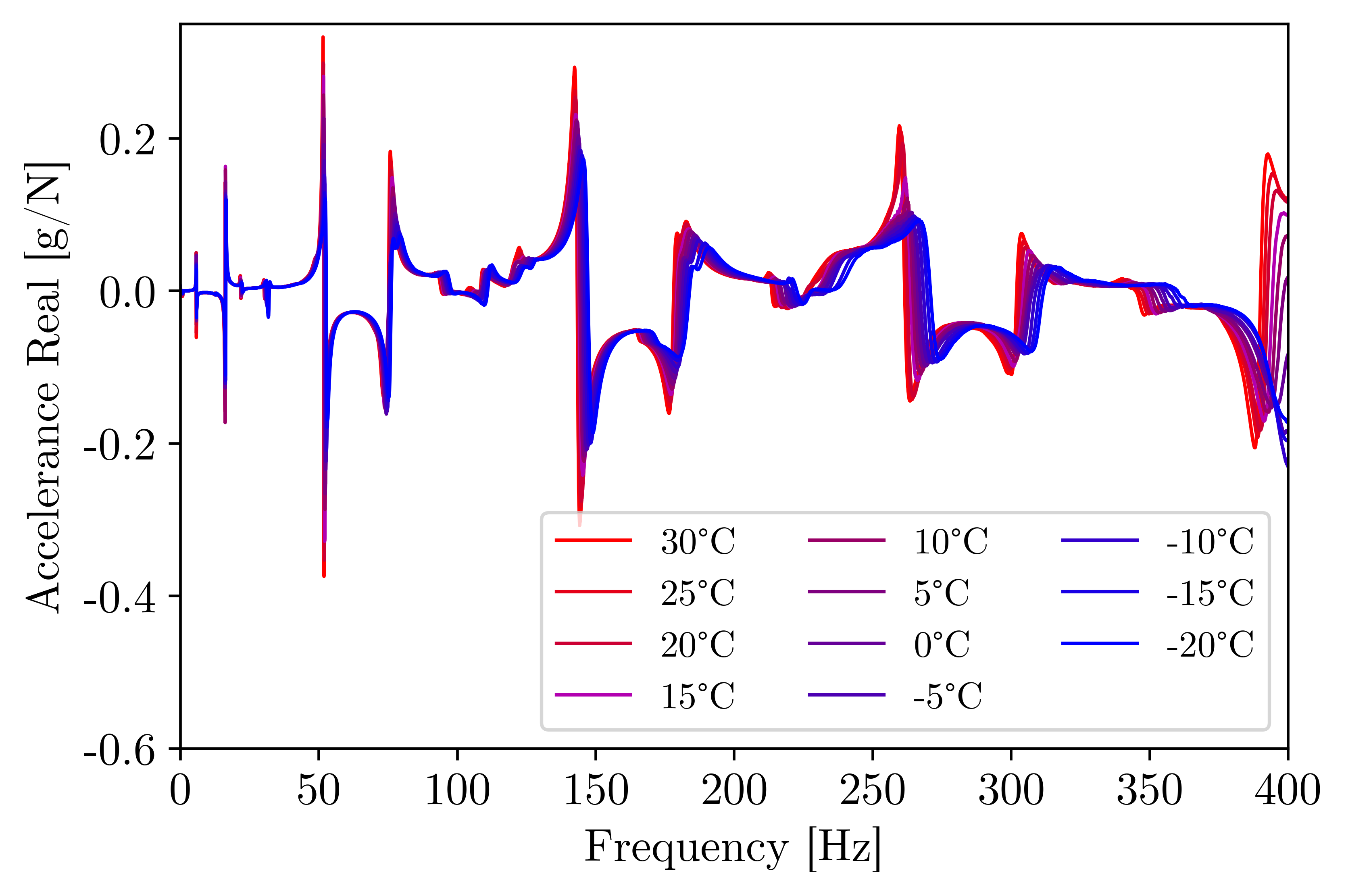}} \\
		\subfloat[\label{fig:FRFs_temps_real_SDOF}]{\includegraphics[width=0.8\columnwidth]{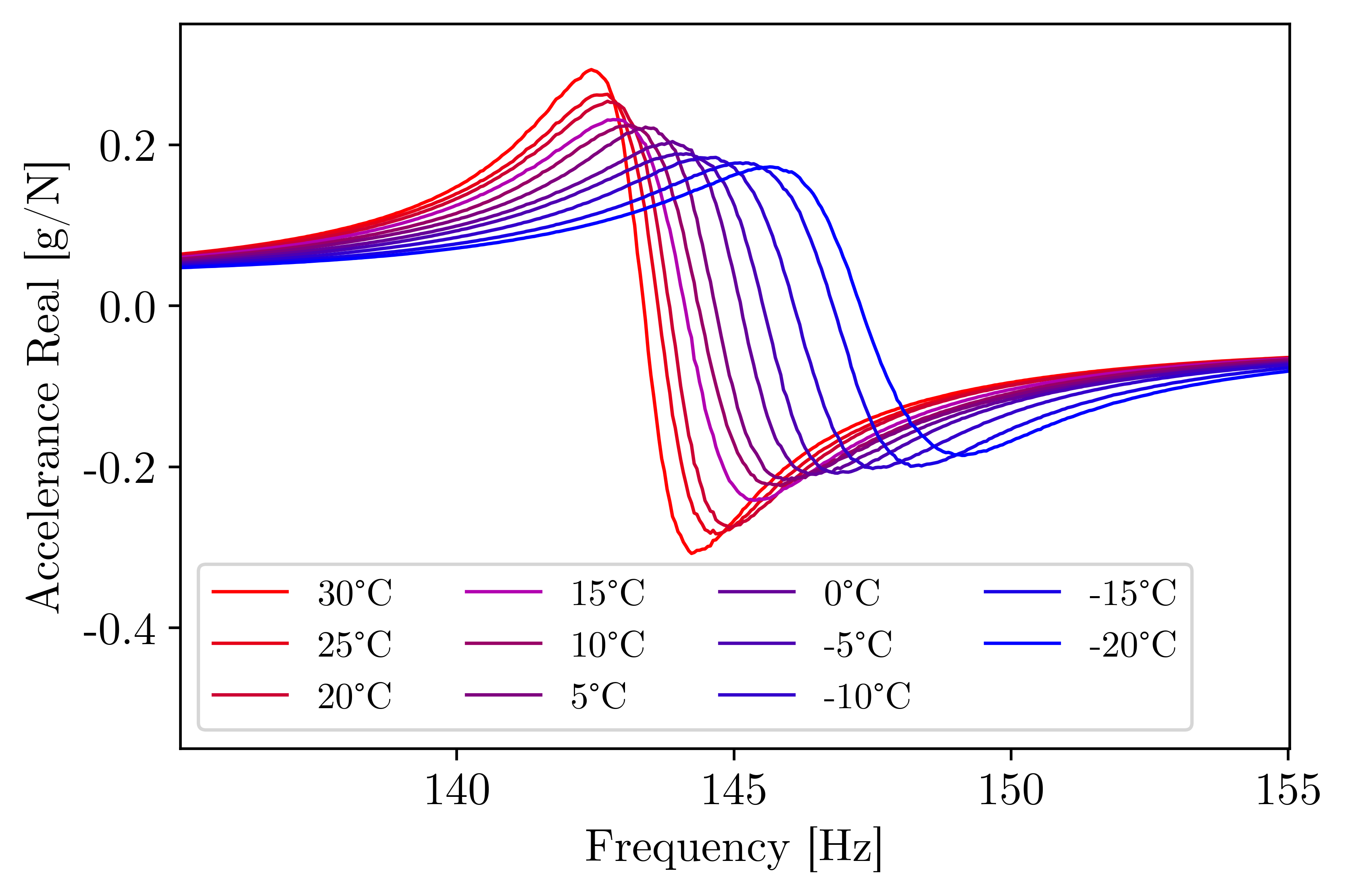}} \\
		\caption{Real components of the helicopter blades FRFs, (a) full bandwidth and (b) bandwidth of interest, at the fourth accelerometer from the blade tip, from vibration data collected at various temperatures in an environmental chamber.}
	\end{figure}

	To minimise the number of hyperparameters in the model and reduce computation time, a narrow frequency band was selected around a lightly-damped, well-separated modal peak. Specifically, the chosen bandwidth was between 135 and 155 Hz, with a higher-order bending mode dominating the response in this band. An SDOF assumption was imposed. (Again, this assumption was considered acceptable for these preliminary analyses, to avoid data-separability issues. Likewise, the coefficients of the functions that relate temperature fluctuations to changes in modal parameters will likely be mode-dependent.) FRFs captured at temperatures -10, -5, 10, and 25$^{\circ}$C provided the training data. The remaining temperature-varied FRFs were used to evaluate the extrapolation results to other temperatures. As with the previous case, the real part was modelled as a probabilistic FRF, using the FRF estimate from Eq.\ (\ref{eq:modalFRFreal}) as the mean of the likelihood function. The real parts of the averaged FRFs for each temperature, from the fourth accelerometer from the blade tip, are shown in Figures \ref{fig:FRFs_temps_real} and \ref{fig:FRFs_temps_real_SDOF}. Figure \ref{fig:FRFs_temps_real} shows the full measured bandwidth, and Figure \ref{fig:FRFs_temps_real_SDOF} shows the FRF in the bandwidth of interest, between 135 and 155 Hz. 
	
	Figures \ref{fig:FRFs_temps_real} and \ref{fig:FRFs_temps_real_SDOF} show a proportional decrease in frequency corresponding to each incremental temperature increase, with discrepancies more noticeable at the higher modes than the lower modes. These results are expected, as the higher-frequency modes are more sensitive to environmental and other changes. The maximum frequency difference among the tests was approximately 15.3 Hz (for modes less than 400 Hz), found in the band between 335 and 375 Hz, as obtained via peak-picking. In the same band, the average frequency difference for each 5$^{\circ}$C increment was approximately 1.5 Hz. For the mode at approximately 145 Hz shown in Figure \ref{fig:FRFs_temps_real_SDOF}, the maximum frequency difference was approximately 3.8 Hz, and the average frequency difference for each 5$^{\circ}$C increment was approximately 0.38 Hz. Further details regarding the data collection and processing for these tests can be found in \cite{DardenoISMA2022}.

\vspace{12pt} 
\section{Case 1: population-based modelling of FRFs for nominally-identical structures}
\label{Case1}

	The first case used FRF data from four nominally-identical helicopter blades, collected at ambient laboratory temperature. A population \emph{form} for the FRFs of the helicopter blades was developed using the aforementioned hierarchical (partial-pooling) approach. Models were developed using the probabilistic programming language \texttt{Stan}. Analyses were performed using MCMC, via the no U-turn (NUTS) implementation of Hamiltonian Monte Carlo (HMC) \cite{hoffman2014no, betancourt2015hamiltonian}. HMC uses approaches based in differential geometry to generate transitions that span the full marginal variance, which allows the algorithm to accommodate large curvature changes in the target distribution (which are common with hierarchical models) \cite{betancourt2015hamiltonian}.
	
	\subsection{Model development} \label{Case1_model}
	The population of helicopter blade FRFs, with frequency inputs, $ \boldsymbol{\omega}_{k} $, and accelerance outputs, $ \mathbf{H}_{k} $, was re-written from the general representation in Eq.\ (\ref{eq:PopulationData_1}),
	
	\begin{equation}
		\left\{\boldsymbol{\omega}_{k},\mathbf{H}_k\right\}_{k=1}^K =
		\left\{\left\{\omega_{ik},H_{ik}\right\}_{i=1}^{N_k}\right\}_{k=1}^K
		\label{eq:PopulationDataFRF_1}
	\end{equation}
	
	\noindent where $ \left\{\omega_{ik},H_{ik}\right\} $ was the $ i^{th} $ pair of observations in domain $ k $. Then, considering only the real component of the FRF, Eq.\ (\ref{eq:PopulationData_2}) was re-written as, 
	
	\begin{equation}
		\left\{\textrm{real}\left[H_{ik}\right] = f_k\left(\omega_{ik}\right) + \epsilon_{ik}\right\}_{k=1}^{K}
		\label{eq:PopulationDataFRF_2}
	\end{equation}
	
	\noindent where for each $ i^{th} $ observation, the output was determined by evaluating one of $ K $ latent functions, $ f_k\left(\omega_{i,k}\right) $, plus additive noise, $ \epsilon_{i,k} $. For this work, there were four helicopter blades and four domains. Therefore, the population model included four latent functions (i.e., $ K = 4 $). To accommodate outliers in the FRF data, the (real) FRFs were modelled probabilistically with an assumed Student's t-distributed likelihood, 
	
	\begin{equation}
		\text{real}\left[\mathbf{H}_k\left(\boldsymbol{\omega}_{k}\right)\right] \sim \mathcal{T}\left(\nu, \mathbf{f}_k\left(\boldsymbol{\omega}_{k}\right),\sigma^2\right)
		\label{eq:Hdist}
	\end{equation}

	\noindent where $ \mathbf{f}_k\left(\boldsymbol{\omega}_{k}\right) $ was equal to the real component of the FRF, calculated using modal parameters via Eq.\ (\ref{eq:modalFRFreal}). For both the independent and partial-pooling models, the degrees-of-freedom parameter, $ \nu $, was assumed to be four (4). Setting $ \nu $ to a constant was necessary to stabilise the independent (no-pooling) models for the data-poor domains, to avoid the sampler falsely identifying the modal peaks as noise. (Note that struggling to fit the $ \nu $ parameter was not a problem for the data-rich independent models or for the partial-pooling model, only for the data-poor independent models.) The additive noise variance, $ \sigma^2 $, of the FRFs was assumed to be the same for each blade. This assumption was reasonable, as the same data acquisition system and sensors were used among the different tests. Note that $ \boldsymbol{\omega}_{k} $ was permitted to vary depending on the given FRF. This allowed for flexibility, to consider FRFs with different numbers of spectral lines or missing data points, and to represent population uncertainty.
	
	Natural frequencies and modal damping were learnt at the domain level, and allowed to vary among the different helicopter blades. Residues were shared among the different domains and learnt at the population level, to mitigate model identifiability issues. Shared, population-level prior distributions (with hyperpriors) were also placed over the modal parameters to capture/infer the similarity among the FRFs. Domain-level natural frequencies, $ \boldsymbol{\omega}_{nat} = \{\{\omega_{nat}^{k,m}\}_{m=1}^2\}_{k=1}^4 $,  were sampled from a truncated normal parent distribution, with higher-level expectation and variance sampled from truncated normal distributions,
	
	\begin{equation}
		\begin{split}
			&	\boldsymbol{\omega}_{nat} \sim \mathcal{TN}\left(\boldsymbol{\mu}_{\omega_{nat}},\boldsymbol{\sigma}^2_{\omega_{nat}}\right) \\
			&	\boldsymbol{\mu}_{\omega_{nat}} \sim \mathcal{TN}\left([190,335],[5^2,5^2]\right) \\
			&   \boldsymbol{\sigma}^2_{\omega_{nat}} \sim \mathcal{TN}\left([5,5],[5^2,5^2]\right) \\ 		 
		\end{split}
		\label{eq:pop_natural_freq}
	\end{equation}

	\noindent Note that hyperpriors for $ \boldsymbol{\omega}_{nat} $ are shown in rad/s. Domain-level damping, $ \boldsymbol{\zeta} = \{\{\zeta_{k,m}\}_{m=1}^2\}_{k=1}^4 $, was sampled from a beta parent distribution, with higher-level shape parameters sampled from truncated normal distributions,

	\begin{equation}
		\begin{split}
			&	\boldsymbol{\zeta} \sim \mathcal{B}\left(\boldsymbol{\alpha}_{\zeta},\boldsymbol{\beta}_{\zeta}\right) \\
			&	\boldsymbol{\alpha}_{\zeta} \sim \mathcal{TN}\left([6,6],[0.5^2,0.5^2]\right) \\
			&   \boldsymbol{\beta}_{\zeta} \sim \mathcal{TN}\left([500,500],[50^2,50^2]\right) \\ 		 
		\end{split}
		\label{eq:pop_damping}
	\end{equation}

	\noindent Likewise, shared modal residues, $ \boldsymbol{A} = \{A_m\}_{m=1}^2 $, were sampled from a normal parent distribution, with higher-level expectation and variance sampled from normal and truncated normal distributions, respectively,
	
	\begin{equation}
		\begin{split}
			&	\boldsymbol{A} \sim \mathcal{TN}\left(\boldsymbol{\mu}_{A},\boldsymbol{\sigma}^2_{A}\right) \\
			&	\boldsymbol{\mu}_{A} \sim \mathcal{N}\left([\text{-}0.004,\text{-}0.004],[0.0035^2,0.0035^2]\right) \\ 
			&	\boldsymbol{\sigma}_{A}^2 \sim \mathcal{TN}\left([0.003,0.003],[0.003^2,0.003^2]\right) \\ 		 
		\end{split}
		\label{eq:pop_residue}
	\end{equation}

	\noindent For simplicity, normal distributions were chosen for most of the parameters (or truncated normal distributions, if the parameter was restricted to be positive), although other distributions could be used. A beta distribution was chosen for damping, as beta distributions have support $ x \in [0,1] $, which is suitable for lightly-damped systems. Hyperpriors were chosen based on the physics that could be interpreted by visual inspection of the training data (e.g., natural frequency), or by fitting the data-rich domains independently (discussed below). 
	
	Shared noise variance, $ \sigma^2 $, was sampled from a half-Cauchy parent distribution, with higher-level scale parameter, $ \gamma^2 $, sampled from a truncated normal distribution,
	
	\begin{equation}
		\begin{split}
			&	\sigma^2 \sim \mathcal{C}\left(0,\gamma^2\right) \\
			&	\gamma^2 \sim \mathcal{TN}\left(0.01,0.05^2\right) \\ 	 
		\end{split}
		\label{eq:noise_variance}
	\end{equation}

\begin{figure}[h]
		\centering
		\begin{tikzpicture}[latent/.style={circle, draw, minimum size=1.22cm}]
			\node[obs] (H) {$H_{i,k}$};
			\node[obs,right=1cm of H] (f) {$\omega_{i,k}$};
			\node[latent,above=1cm of H,xshift=1.5cm] (z) {$\boldsymbol{\zeta}_{k}$}; 
			\node[latent,above=1cm of H,xshift=-1.5cm] (w) {$\boldsymbol{\omega}_{nat}^k$}; 
			\node[latent,right=1.75cm of z] (A) {$\boldsymbol{A}$}; 
			\node[latent,above=1cm of w,xshift=-0.75cm] (mu_w) {$\boldsymbol{\mu}_{\omega_{nat}}$}; 
			\node[latent,above=1cm of w,xshift=0.75cm] (sig_w) {$\boldsymbol{\sigma}^2_{\omega_{nat}}$}; 
			\node[latent,above=1cm of z,xshift=-0.75cm] (alpha_z) {$\boldsymbol{\alpha}_{\zeta}$}; 
			\node[latent,above=1cm of z,xshift=0.75cm] (beta_z) {$\boldsymbol{\beta}_{\zeta}$}; 
			\node[latent,above=1cm of A,xshift=-0.75cm] (mu_A) {$\boldsymbol{\mu}_{A}$}; 
			\node[latent,above=1cm of A,xshift=0.75cm] (sig_A) {$\boldsymbol{\sigma}^2_{A}$};  
			\node[latent,left=3cm of H,yshift=1cm] (sig_H) {$\sigma^2$}; 
			\node[latent,above=1cm of sig_H] (gamma_sig) {$\gamma^2$}; 
			\node[const,below=1cm of sig_H] (nu) {$\text{ }\nu \text{ }$}; 
			\plate [inner sep=.25cm,yshift=.1cm,xshift=-.1cm] {plateN} {(H)(f)} {$i \in 1:N_k$};
			\plate [inner sep=.25cm,yshift=.1cm,xshift=0cm] {plateK} {(plateN)(w)(z)} {$k \in 1:K$};
			\edge {w,A,z,f,sig_H} {H} 
			\edge {mu_w,sig_w} {w} 
			\edge {alpha_z,beta_z} {z}
			\edge {mu_A,sig_A} {A}
		\edge {gamma_sig} {sig_H}
		\edge {nu} {H}
		\end{tikzpicture}
		\caption{Graphical representation of the hierarchical FRF model developed for the first case.}
		\label{fig:DGM_ambient}
	\end{figure}
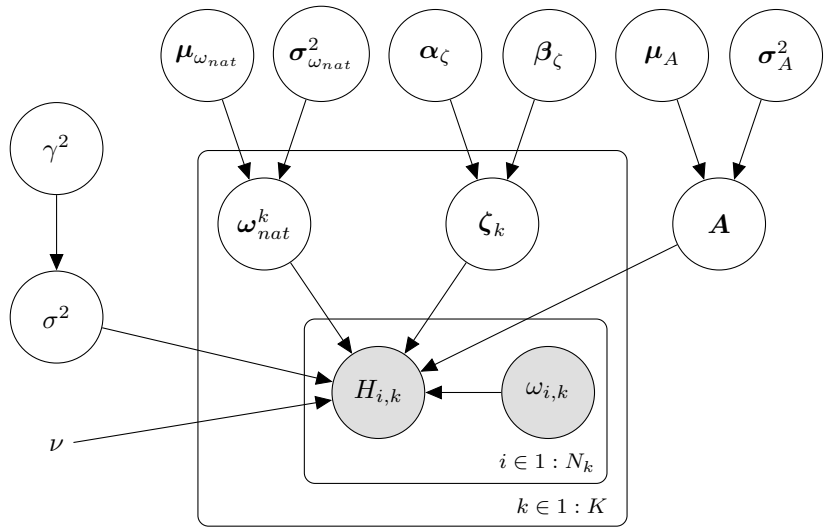

	\noindent The half-Cauchy was considered an appropriate prior distribution for the shared noise term, as it peaks at zero, but has a long right tail, to allow for some flexibility in the assumptions \cite{gelman2006prior}. (Note that, in practice, measurement noise would be expected to be fairly small given that FRFs are typically averaged.) A graphical model displaying the parameter hierarchy is shown in Figure \ref{fig:DGM_ambient}.
	
	The models were trained using data with added noise, to avoid over-fitting small deviations/lightly-participating modes within the band of interest. This was accomplished by adding zero-mean Gaussian noise, with variance equal to 10\% of the absolute maximum amplitude of the data from Blade 1, to the 20 time-data blocks prior to computing FRFs. Then, the time data were subsampled such that there were two data-rich domains (Blades 1 and 2), and two data-poor domains (Blades 3 and 4). (Note that the data-rich groups were subsampled as well. This procedure was adopted to reduce the dimensionality of the training data, although there were enough training points to characterise each mode within the band of interest for these blades.) FRFs were then computed and averaged, and truncated to the desired bandwidth. This procedure resulted in 281, 207, 11, and 31 training points for Blades 1, 2, 3, and 4, respectively. 

	The intent was that the two data-rich FRFs would lend statistical support to the sparser domains, thereby reducing uncertainty compared to an approach without pooling. (Note that partial pooling is also applicable to other situations, such as having only a single data-rich member, as briefly explored in \ref{appendix}.) Recall the grouping of the peaks from the FRFs shown in Figures \ref{fig:FRFs_allblades_real} and \ref{fig:FRFs_allblades_real_2DOF}. Both data-rich FRFs belonged to one of the groupings; likewise, the two data-poor FRFs belonged to the other grouping. This represents a challenging situation where the model can be heavily biased towards the available training data. The hierarchical/partial-pooling modelling structure allows the target distributions to be informed by the data (for very data-poor domains, this will of course be limited), but with the help of the full population dataset (including the data-rich domains). In addition, the subsampling and added noise resulted in some outliers in the data, creating a challenging inference problem.
	
	The \texttt{Stan} HMC sampler was run using four chains, with a target average proposal acceptance probability rate of 0.8 (this parameter controls the target acceptance rate of the NUTS algorithm, and setting it to a value close to 1 can reduce the number of false-positive divergences \cite{Standivergences}). The sampler was run for 10000 samples per chain (and an additional 5000 warm-up samples were discarded per chain to diminish the influence of the starting values \cite{gelman2013bayesian}), for each parameter. Models were also run for each set of blade data independently (no pooling), for comparison with the partial-pooling model. Posterior predictive checks were performed to ensure stationarity and proper mixing of the Markov chains. As computed by \texttt{Stan}, the \emph{maximum} split $ R_{hat} $ diagnostic, and the \emph{minimum} effective sample size ($ N_{Eff} $), considering all parameters/hyperparameters, are shown in Table \ref{tab:MCMCchecks_Case1}, for all models. Note that split $ R_{hat} $ values that do not exceed 1.01 suggest proper mixing of the chains \cite{Vehtari2021}. Likewise, total $ N_{Eff} $ values greater than 400 are considered suitable for stable estimates and proper interpretation of split $ R_{hat} $, for a setup comprised of four parallel chains, as employed in this work \cite{Vehtari2021}. These criteria were met for all models developed.
	
		\begin{table}[h]
		\renewcommand{\arraystretch}{1.5}
		\caption{\label{tab:MCMCchecks_Case1}Posterior predictive diagnostics for Case 1.}
		\begin{center}
			\begin{tabular}{ p{1.55cm} p{2.6cm} p{1.2cm} p{1.2cm} p{1.2cm} p{1.2cm}}
				\hline
				{} & Partial Pooling & \multicolumn{4}{c}{No Pooling} \\
				\hline
				Metric & All Blades & Blade 1 & Blade 2 & Blade 3 & Blade 4 \\
				\hline
				min $ N_{Eff} $ & 3085 & 518 & 7583 & 1290 & 1251 \\
				max $ R_{hat} $  & 1 & 1 & 1 & 1 & 1 \\
				\hline
			\end{tabular}
		\end{center}
	\end{table}
	
	\subsection{Model results} \label{Case1_results}
\begin{figure}[h!]
		\centering
		\subfloat[\label{fig:B1_both_models}]{\includegraphics[width=0.8\textwidth]{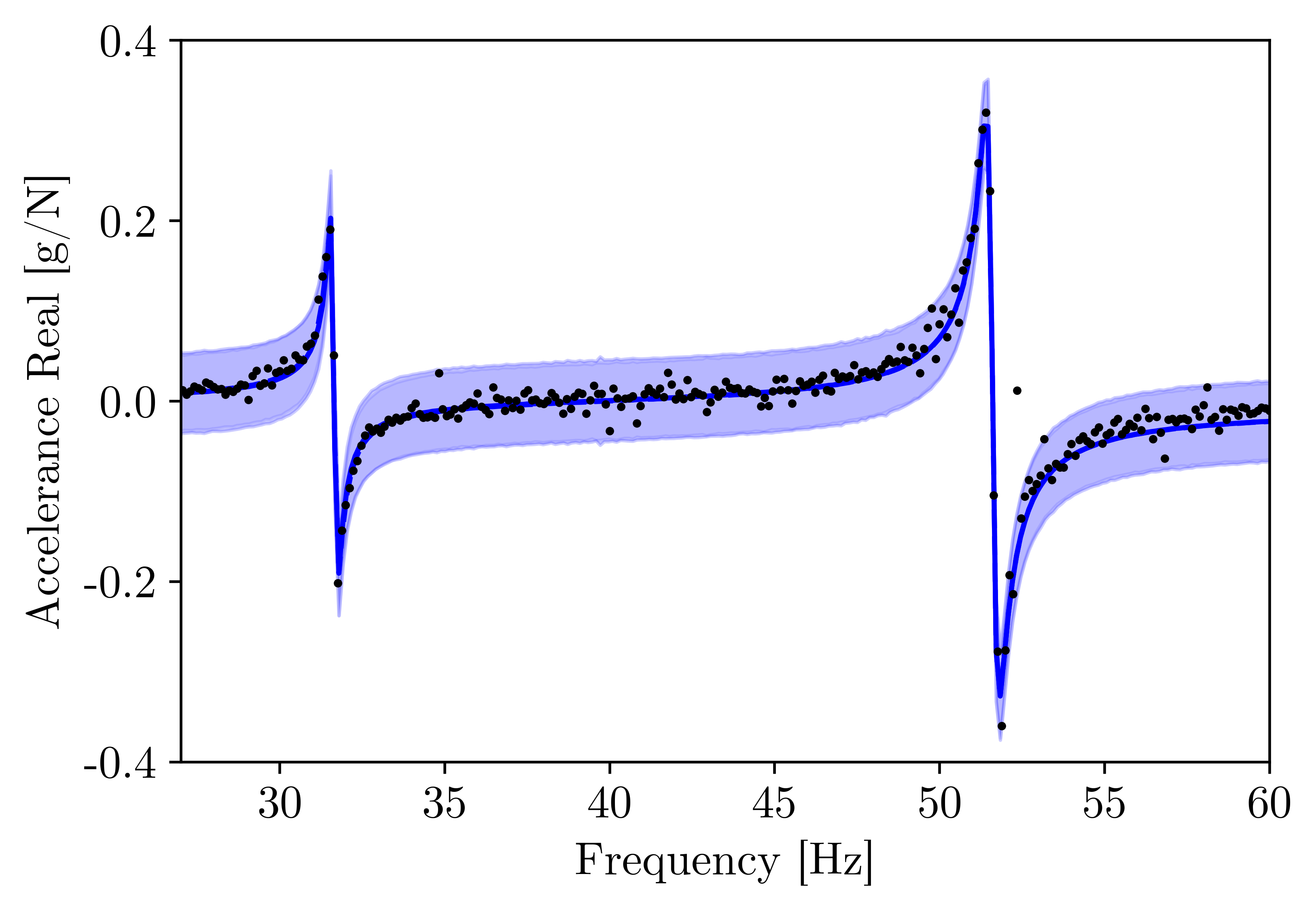}} \\
		\subfloat[\label{fig:B2_both_models}]{\includegraphics[width=0.8\textwidth]{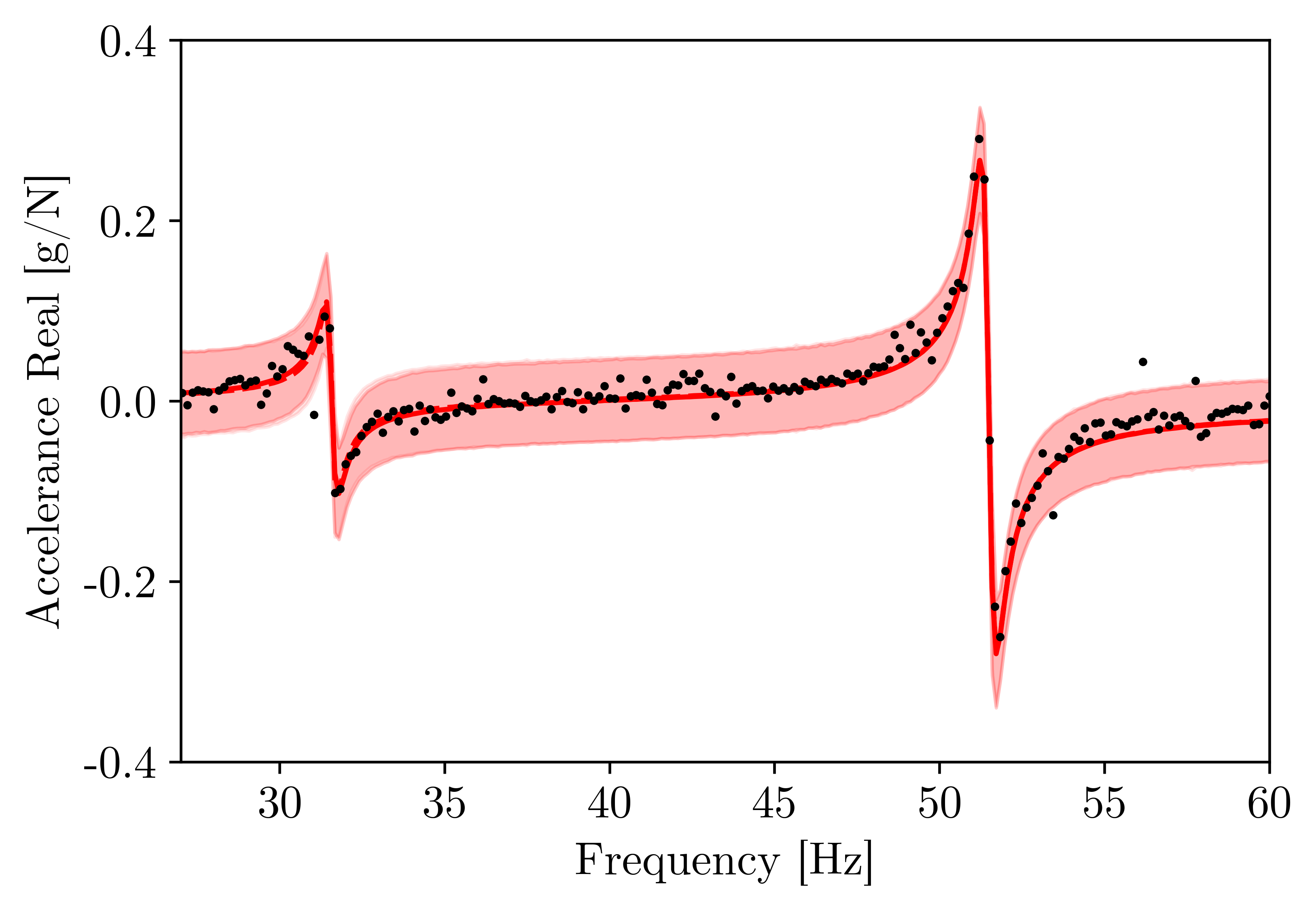}} \\
		\caption{Posterior predictive mean and 3-sigma deviation for independent (no-pooling) and partial-pooling models, for data-rich domains (a) Blade 1 (blue), and (b) Blade 2 (red). Training data are shown in a black scatter plot. Solid lines are the posterior predictive means for the partial-pooling models, and dashed lines are the posterior predictive means for the independent models. The variance is represented by shaded regions, where the independent model variances are shown in a lighter colour than those for the partial-pooling models.}
	\end{figure}

\begin{figure}[h!]
		\centering
		\subfloat[\label{fig:B3_both_models}]{\includegraphics[width=0.8\textwidth]{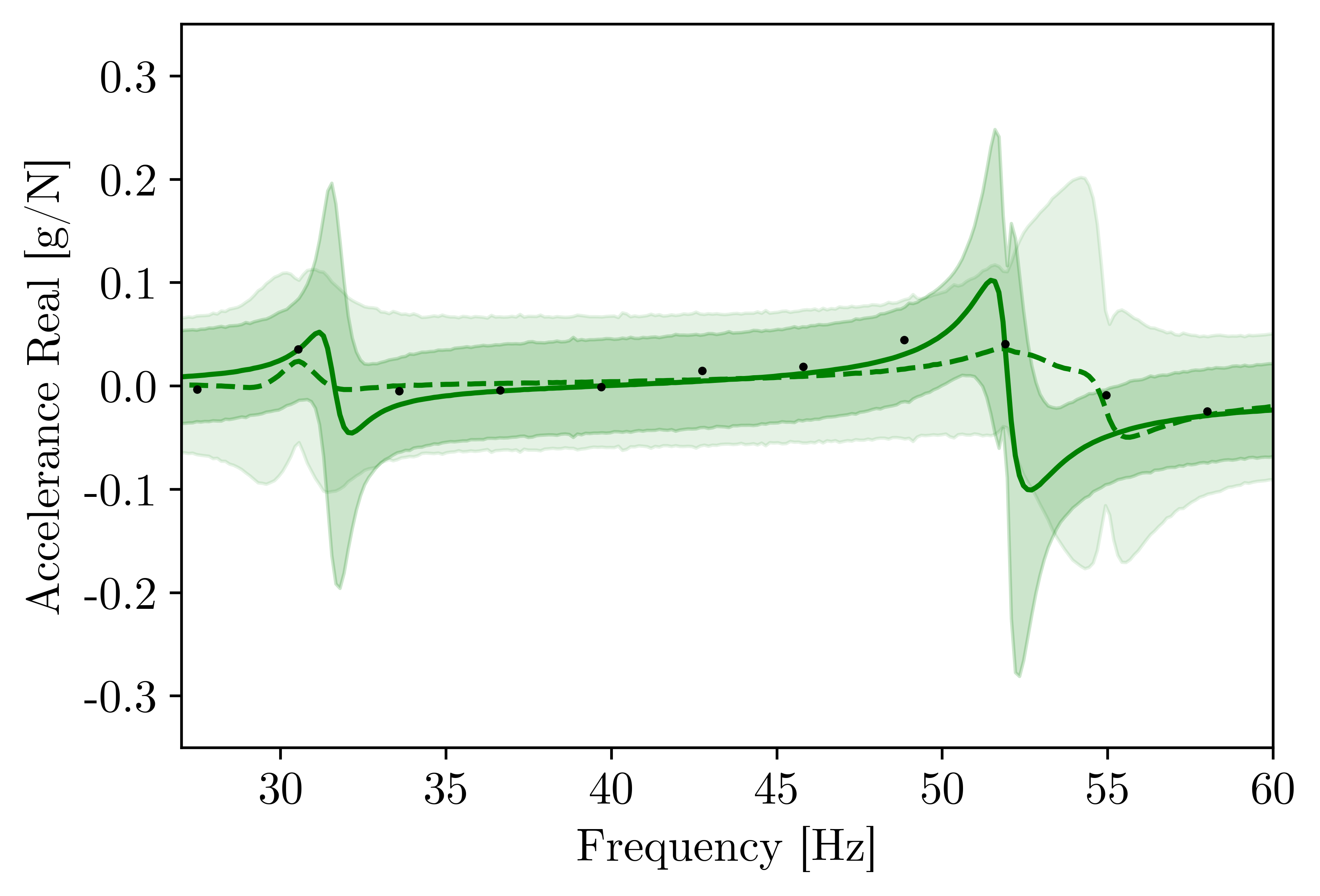}} \\
		\subfloat[\label{fig:B4_both_models}]{\includegraphics[width=0.8\textwidth]{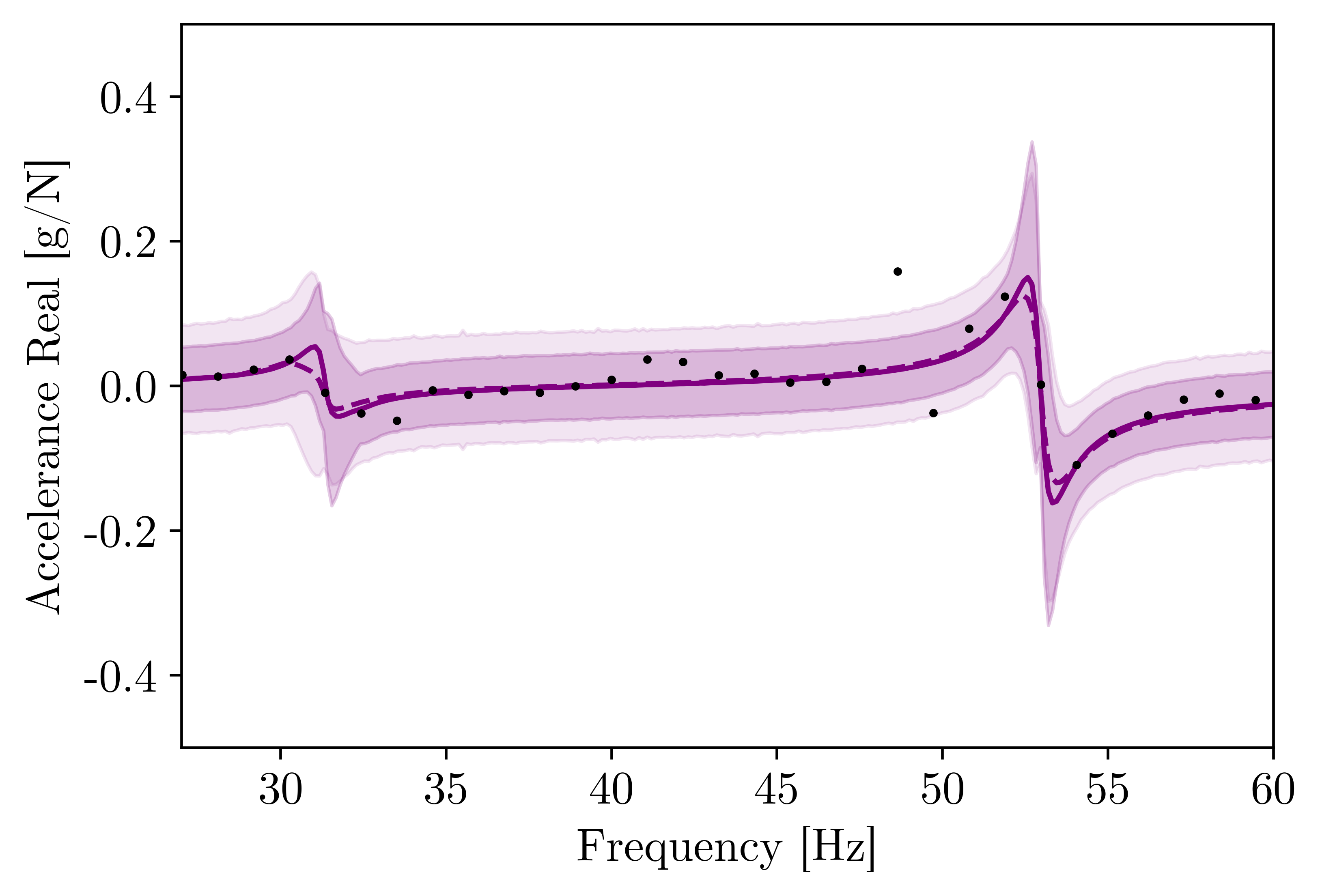}} \\
		\caption{Posterior predictive mean and 3-sigma deviation for independent (no-pooling) and partial-pooling models, for data-poor domains (a) Blade 3 (green) and (b) Blade 4 (purple). Training data are shown in a black scatter plot. Solid lines are the posterior predictive means for the partial-pooling models, and dashed lines are the posterior predictive means for the independent models. The variance is represented by shaded regions, where the independent model variances are shown in a lighter colour than those for the partial-pooling models.}
	\end{figure}

	Eq.\ (\ref{eq:modalFRFreal}) was used to compute FRFs from the posterior MCMC samples of the modal parameters, for each blade. FRFs \emph{with predicted variance} were generated by sampling from a Student's t-distribution with four degrees-of-freedom, mean equal to the computed FRFs, and posterior predictive measurement noise, ($ \sigma^2 $), for each sample. Total variance was estimated by taking the standard deviation of the FRFs. Posterior predictive mean and 3-sigma deviation for the partial-pooling and independent models are plotted in Figures \ref{fig:B1_both_models} to \ref{fig:B4_both_models}, respectively.
	
	Indeed, Figures \ref{fig:B1_both_models} to \ref{fig:B4_both_models} show that the sparser FRFs improved significantly for the combined model, compared to the independent models. With only 11 data points, Blade 3, shown in Figure \ref{fig:B3_both_models} in green, was missing most of the information necessary to characterise the two modes. The independent model for Blade 3 relied heavily on the user-specified prior, while the partial-pooling model borrowed information from the other three blades to inform the shared latent model. Similar (albeit less severe), results were seen for Blade 4, shown in Figure \ref{fig:B4_both_models} in purple. With 31 data points, the FRF for Blade 4 was less sparse than that for Blade 3, but was still somewhat lacking near the resonance peaks, especially for the first mode. Figure \ref{fig:B4_both_models} shows a significant reduction in variance with the partial-pooling approach. The improvements in the data-rich FRFs, Blades 1 (blue) and 2 (red) were not as apparent, as expected. 
	
\begin{figure}[H]
		\centering
		\subfloat[\label{fig:marginal_omega_1}]{\includegraphics[width=0.5\textwidth]{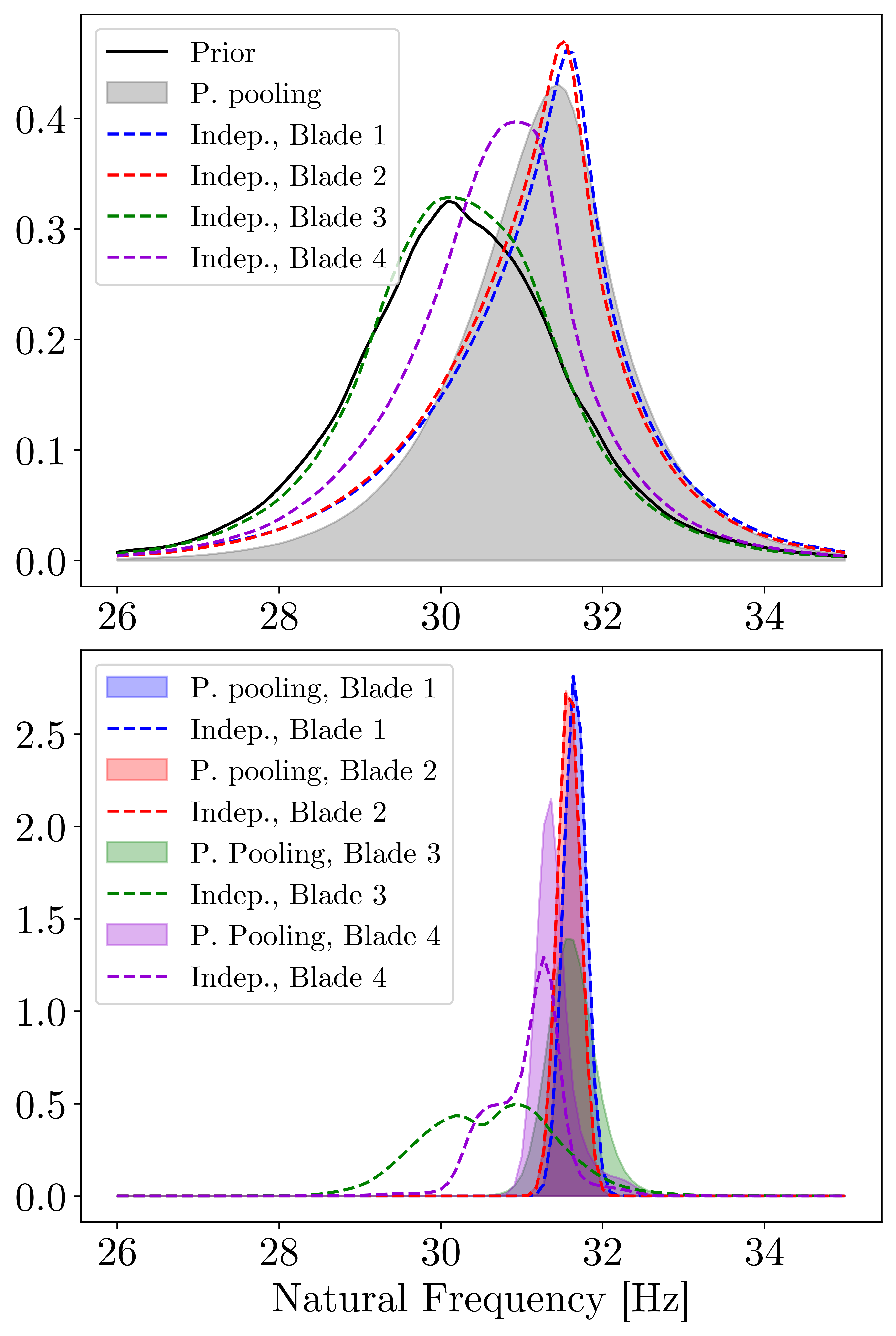}} 
		\subfloat[\label{fig:marginal_omega_2}]{\includegraphics[width=0.5\textwidth]{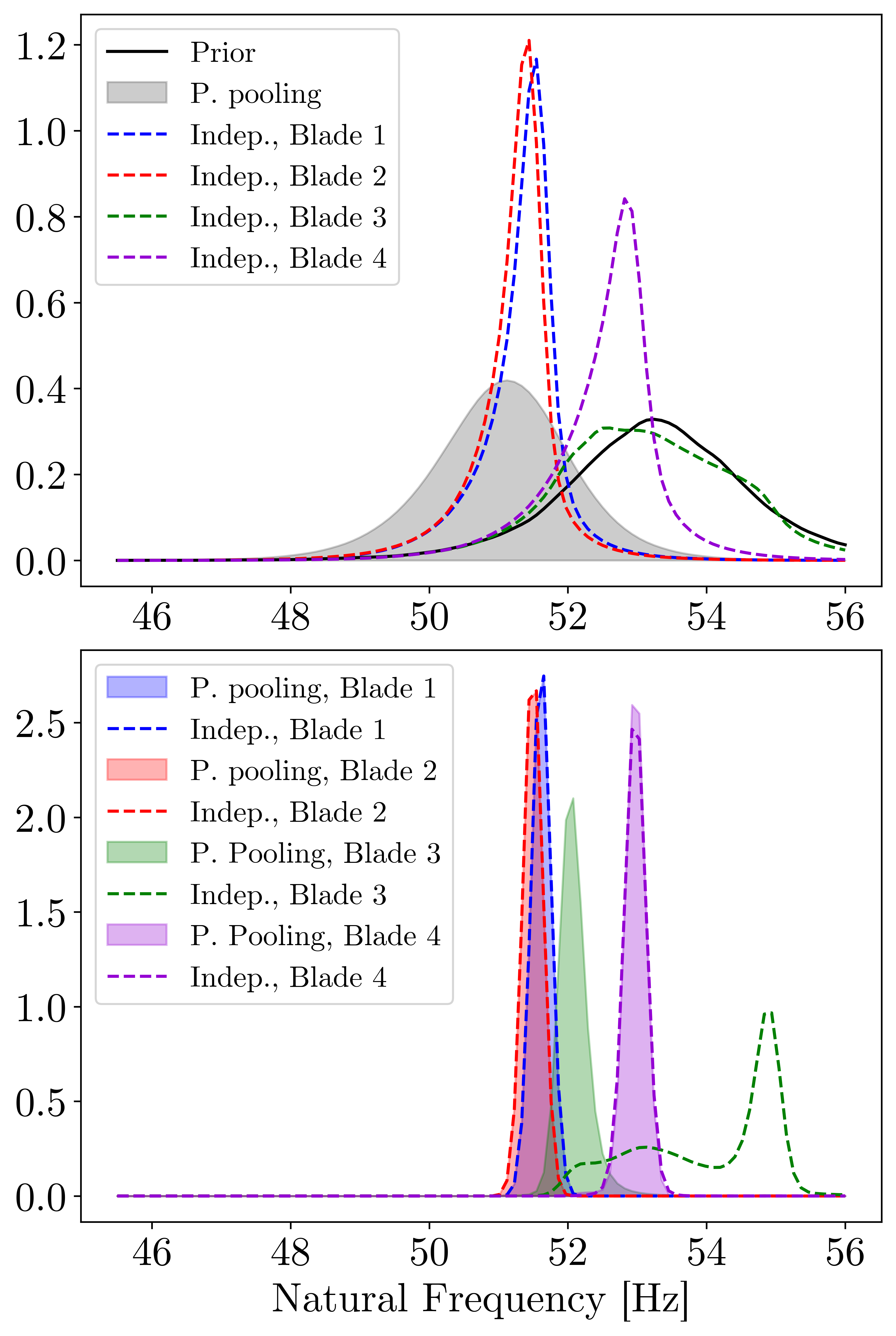}} \\
		\subfloat[\label{fig:marginal_zeta_1}]{\includegraphics[width=0.5\textwidth]{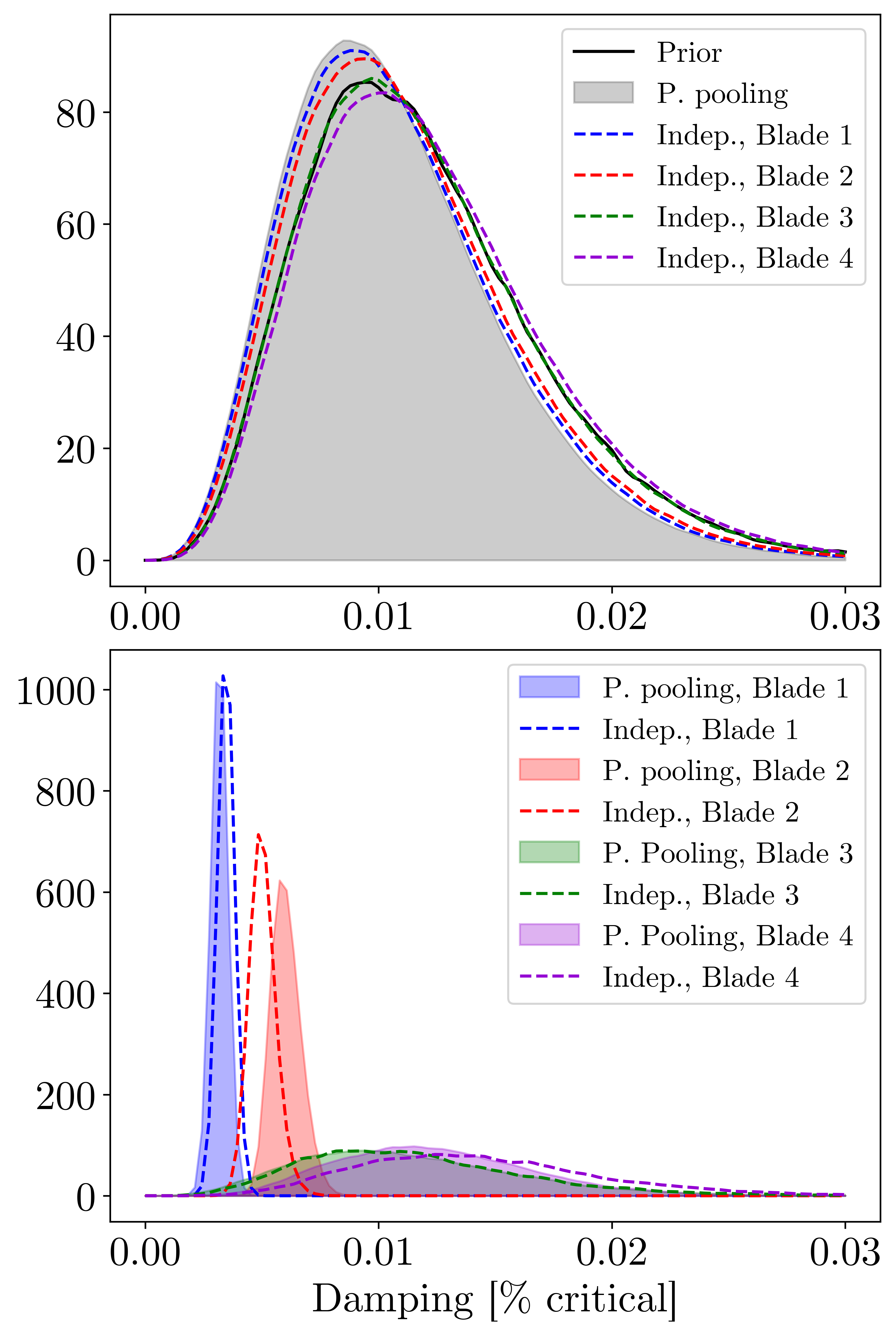}} 
		\subfloat[\label{fig:marginal_zeta_2}]{\includegraphics[width=0.5\textwidth]{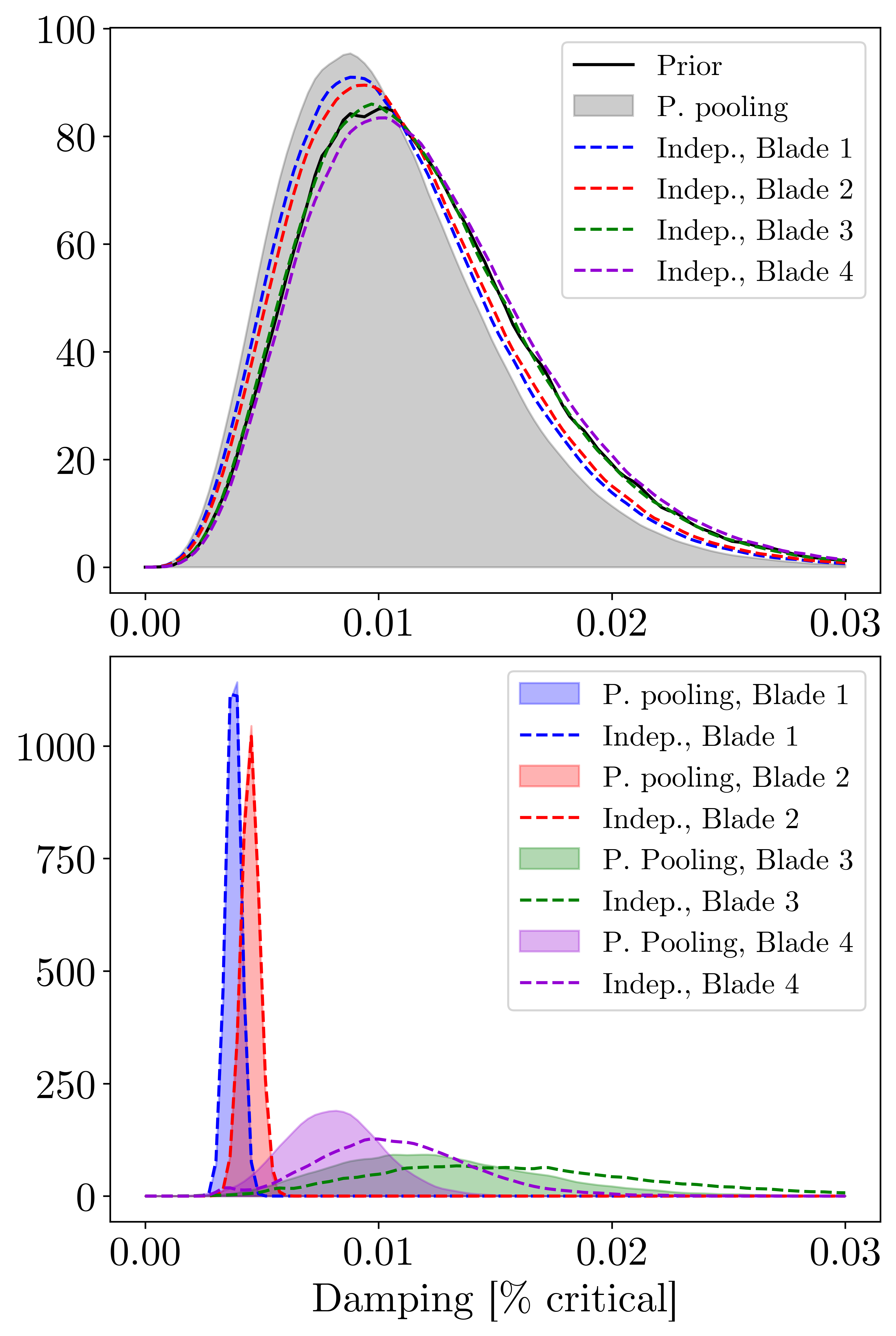}} \\
		\caption{Marginal distributions for natural frequency, (a) mode 1 and (b) mode 2; and for damping, (c) mode 1 and (d) mode 2. Population- and domain-level distributions are shown at the top and bottom, respectively. For the partial-pooling model, posterior distributions are shaded. For the independent (no-pooling) models, posterior distributions are shown as dashed lines. Priors are shown as solid black lines.}
	\end{figure}

\begin{figure}
		\centering
		\subfloat[\label{fig:marginal_A_1}]{\includegraphics[width=0.5\textwidth]{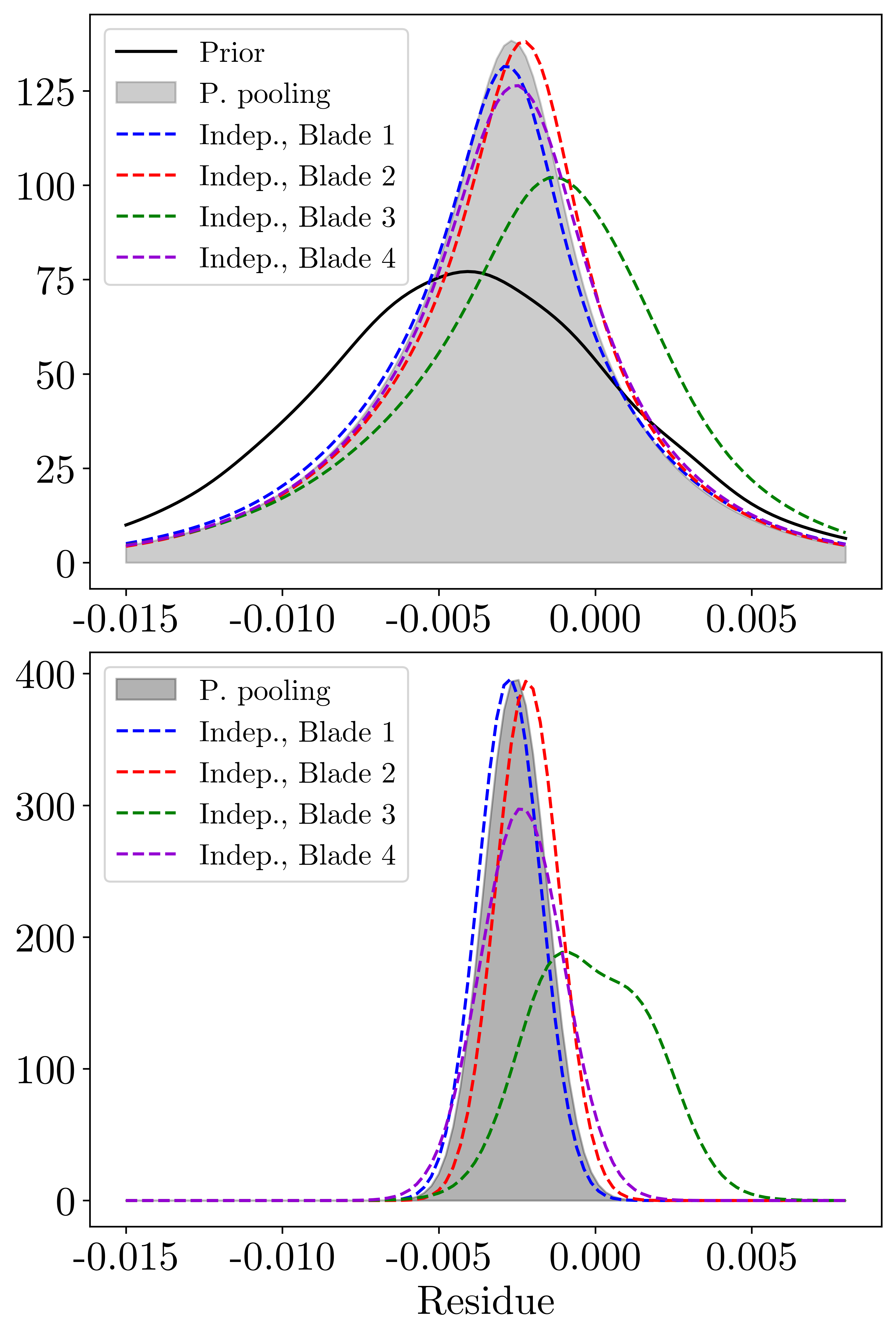}} 
		\subfloat[\label{fig:marginal_A_2}]{\includegraphics[width=0.5\textwidth]{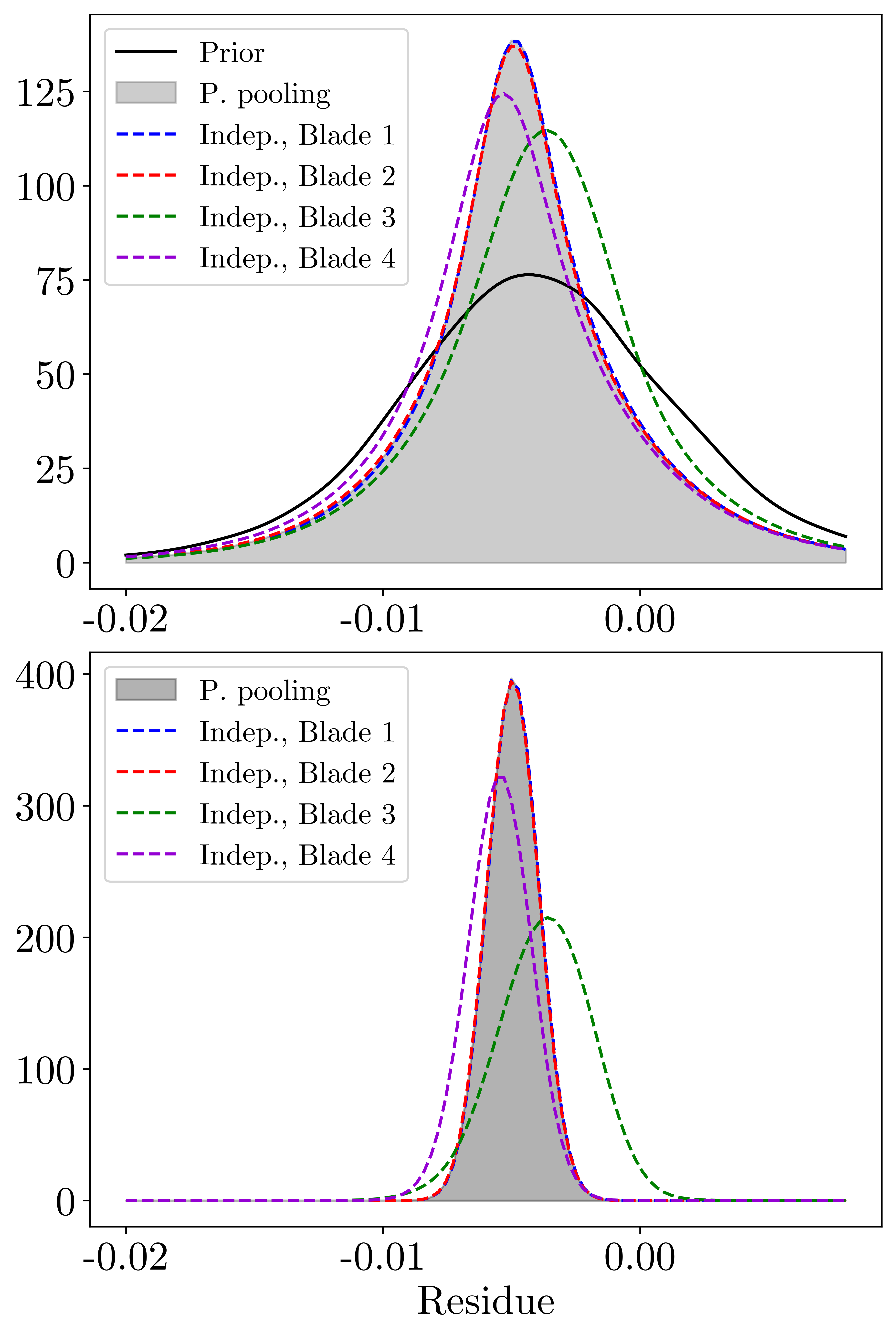}} \\
		\caption{Marginal distributions for residue, (a) mode 1 and (b) mode 2. Parent- and lower-level distributions are shown at the top and bottom, respectively. For the partial-pooling model, posterior distributions are shaded. For the independent (no-pooling) models, posterior distributions are shown as dashed lines. Priors are shown as solid black lines.}
	\end{figure}

	Variance reduction from the combined-inference (partial-pooling) approach can also be visualised by plotting the marginal distributions of the parameters/hyperparameters. For the population-level variables, marginal distributions were approximated by taking the posterior MCMC samples for expectation and variance for each variable (or shape parameters, for damping), and then drawing from a distribution (normal for natural frequency and residue, and beta for damping) with the same statistics as the parameter draws. Kernel density estimation (KDE) was then used to approximate marginal distributions using the population-level draws. (Note it is known what distribution a given sample from the posterior will take, e.g., expectation and variance are learnt assuming \emph{a priori} a normally-distributed mode shape. However, the shape of the posterior distribution is itself unknown, i.e., the distribution over these distributions. Because the posterior is itself a random variable, KDE can be used as an approximation for the marginal distributions, using the posterior MCMC samples.) Marginal distributions for the domain-level variables were also approximated via KDE of the posterior MCMC samples. For the natural frequencies, the population- and domain-level distributions for each mode are shown in Figures \ref{fig:marginal_omega_1} and \ref{fig:marginal_omega_2}. For damping, the population- and domain-level distributions for each mode are shown in Figures \ref{fig:marginal_zeta_1} and \ref{fig:marginal_zeta_2}. For the (shared) residues, which were learnt at the population-level, the parent- and lower-level distributions for each mode are shown in Figures \ref{fig:marginal_A_1} and \ref{fig:marginal_A_2}.
	
	Figures \ref{fig:marginal_omega_1} to \ref{fig:marginal_A_2} show that when the domains were able to share information via common higher-level distributions, the variance was reduced compared to a no-pooling approach. In these figures, the population-level distributions are shown at the top, while domain-level distributions are shown on the bottom of the plots. In Figure \ref{fig:marginal_omega_1}, for the first natural frequency, the population-level distribution of the partial-pooling model was closely aligned with those of the independent models for the data-rich domains, and was taller and with lower variance than those of the independent models for the data-poor domains. This observation was likely because the data-rich domains dominated the available data for that mode, compared to the data-poor domains (forcing the distribution towards the natural frequencies of the data-rich domains). This data sharing was also evident at the domain-level, where the KDEs of the data-poor domains aligned closely with those from the data-rich domains, for the partial-pooling models. Also, in Figure \ref{fig:marginal_omega_1}, it can be seen that the no-pooling KDEs of the data-poor FRFs were primarily informed by the priors. Indeed, the population-level distribution for the sparsest domain, Blade 3, was nearly indistinguishable from the prior, indicating that the data had little influence on the posterior (this makes sense, as there were very little data to characterise this mode). In contrast, Figure \ref{fig:marginal_omega_2} shows that for the second natural frequency, the population-level distribution for the partial-pooling model was flatter than those for the independent models. This was likely to have occurred because more data were available for the second peak from Blade 4, which was from the second grouping (widening the distribution to accommodate the greater differences in natural frequency). Again, this conclusion is supported at the domain-level, where the KDE for Blade 4 is very similar for the partial-pooling and independent models (although there is still a small variance reduction for the partial-pooling model). For Blade 3, which was quite data-poor at the second mode as well as the first, the KDE in Figure \ref{fig:marginal_omega_2} shows that the natural frequency was estimated to occur in between those of the other, more data-rich blades. Again, for Blade 3, the no-pooling KDE differed significantly from the others, as this parameter was primarily informed by the priors, because of a lack of data at the second mode. For both modes, it is clear that the independent models struggled to fit the data-poor FRFs, as evidenced by bimodal distributions for natural frequency. 
	
	Figures \ref{fig:marginal_zeta_1} and \ref{fig:marginal_zeta_2} show that the population-level distributions for damping did not vary significantly among the partial-pooling and independent models (with only a slight deviation from the priors for the partial-pooling and data-rich models), likely because the priors were fairly strong and also accurate to the data. At the domain-level, KDEs for the data-rich FRFs showed high similarity for the different pooling models, while KDEs for the data-poor FRFs tended towards the priors. Similar results were seen for the residues, in Figures \ref{fig:marginal_A_1} and  \ref{fig:marginal_A_2}, except that the residues were shared among the domains for the partial-pooling model, so there was only one lower-level distribution for each mode. Close alignment of the domain-level distributions, of the data-rich domains, suggests that sharing the residue among the different blades was a suitable assumption.
	
	The results presented in this case demonstrate the development of a population form, where a combined hierarchical FRF model is learnt for a group of nominally-identical helicopter blades. Two domains (Blades 3 and 4) had limited data, and were especially sparse near the resonance peaks. Such a situation could occur if an insufficient sampling rate was selected during the data acquisition process. By borrowing data from data-rich domains within the population, variance was reduced compared to an independent modelling approach.

\vspace{12pt} 
\section{Case 2: population-based modelling of FRFs with temperature variation}
\label{Case2}
	The second case also used a hierarchical modelling structure with partial pooling, but assumed non-exchangeable models whereby parameters of the FRF varied with respect to temperature, i.e., the parameters were conditioned on the data. The goal was to learn functional relationships (at the population level), between temperature and natural frequency, and between temperature and damping, so that inferences could be made at temperatures not used in model training. As with the first case, all models in the second case were developed using \texttt{Stan}, and analyses were performed using MCMC, via the no U-turn (NUTS) implementation of Hamiltonian Monte Carlo (HMC) \cite{hoffman2014no, betancourt2015hamiltonian}. 
	
	\subsection{Identification of appropriate functional relationships between temperature and the modal parameters} \label{Case2_functions}
\begin{figure}[h]
		\centering
		\subfloat[\label{fig:wn_least_squares}]{\includegraphics[width=0.5\textwidth]{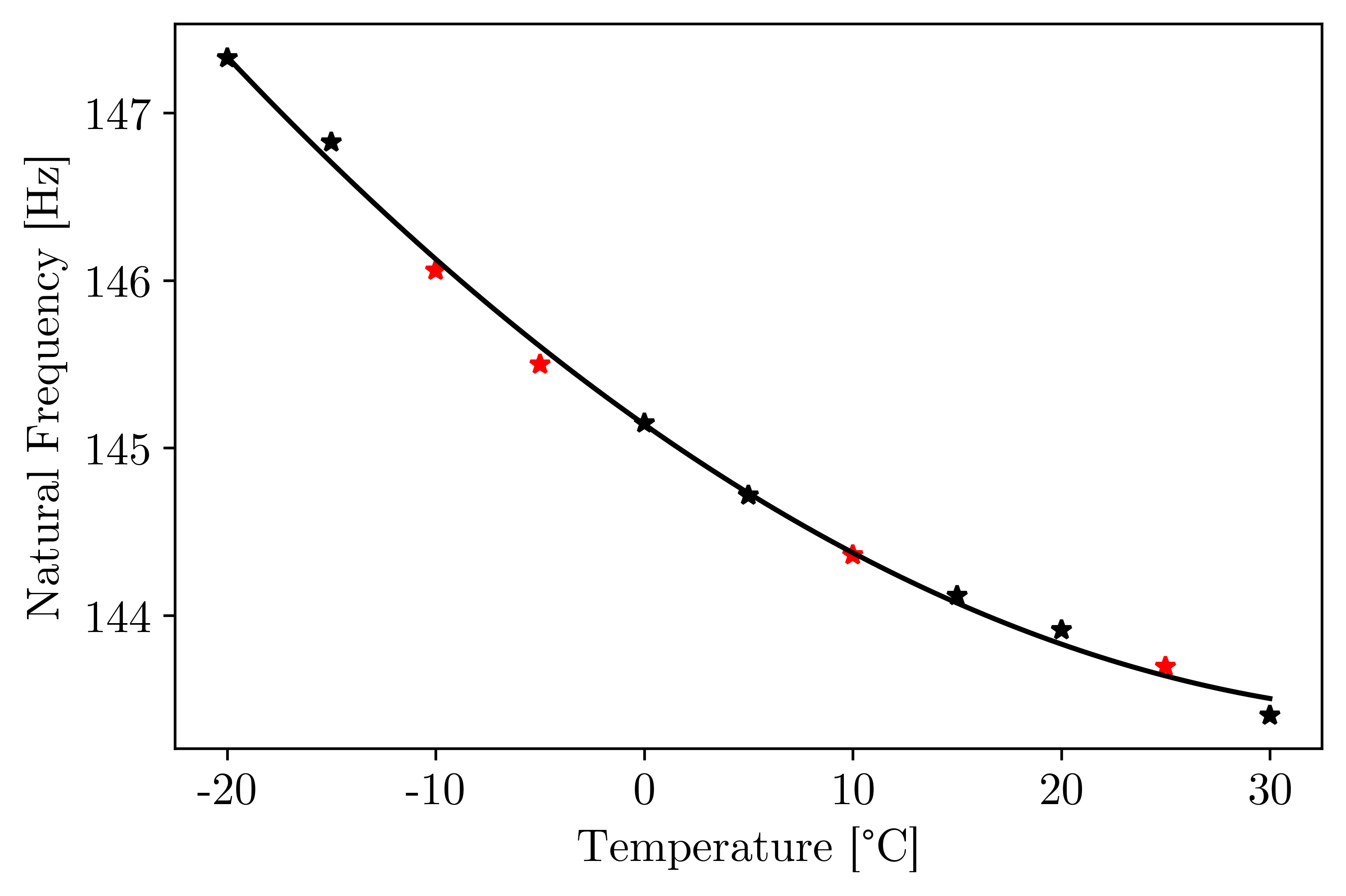}} 
		\subfloat[\label{fig:zeta_least_squares}]{\includegraphics[width=0.5\textwidth]{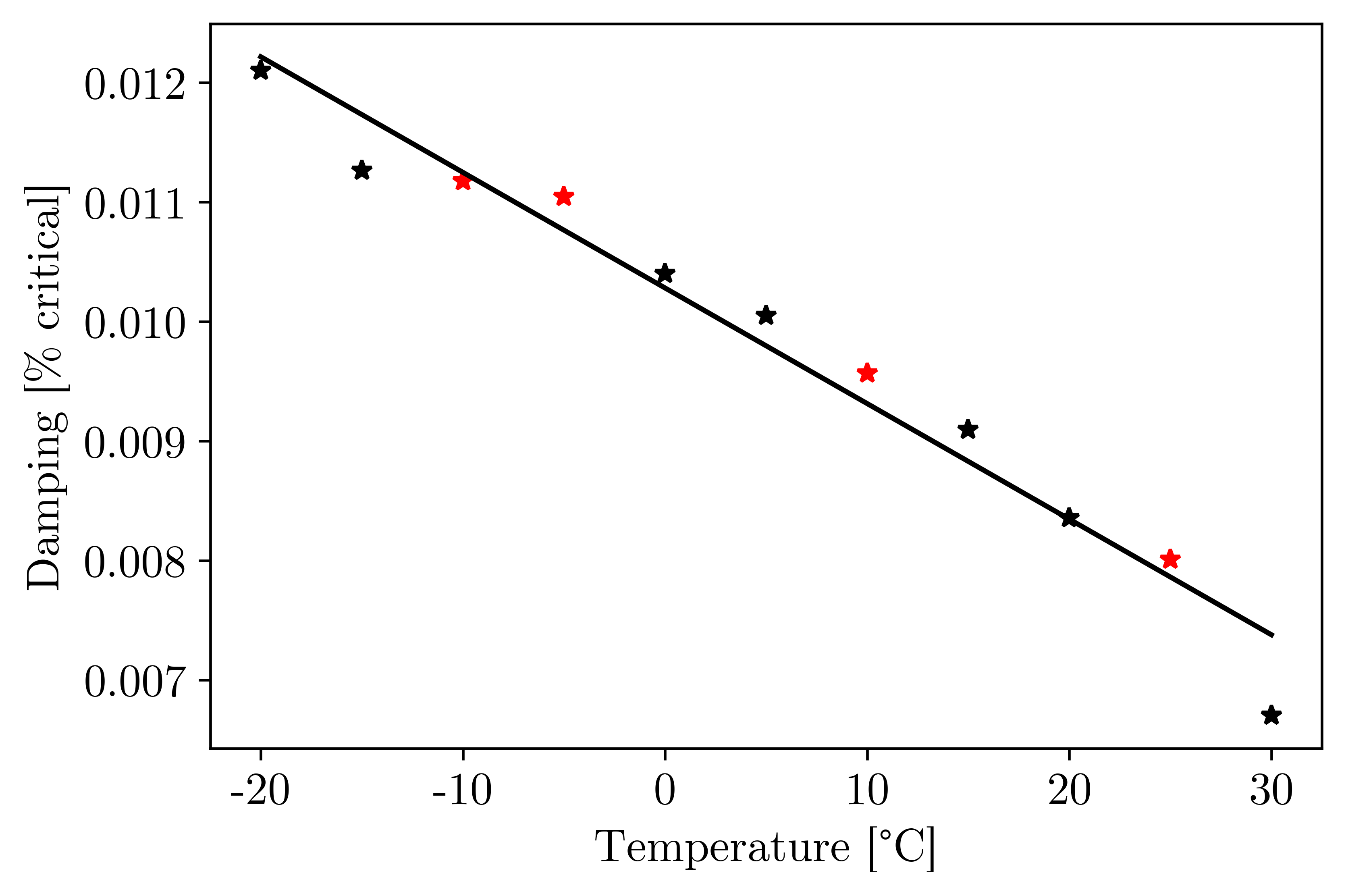}} 
		\caption{Ground-truth estimates for (a) natural frequency and (b) damping, plotted against temperature. Second-order polynomial (for natural frequency) and linear (for damping) least-squares fits are shown as solid black lines. Parameters associated with FRFs used to train the model (-10, -5, 10, and 25$^{\circ}$C) are shown in red.}
	\end{figure}

	Prior to developing the partial-pooling model, it was useful to determine appropriate functions for the temperature relationships. Independent models for each (real) temperature-varied FRF from Figure \ref{fig:FRFs_temps_real_SDOF} were fitted, and ground-truth values were estimated by computing the expectation for each modal parameter. (Note that in the second case, the modal parameters from the independent models for FRFs at all measured temperatures were considered to be the ground truth, as each FRF was data-rich. This process of using the same data to train and develop the model structure is called post-selection inference \cite{LeeSun2016}, and must be used cautiously, as models that employ the technique are at risk of overfitting \cite{LB_MRJ,gelman2013bayesian}. However, it was assumed for this work that because the number of candidate models was small, the bias resulting from using the data twice was also small \cite{LB_MRJ,gelman2013bayesian}.) The ground-truth estimates for natural frequency and damping are plotted against temperature, and shown with least-squares fits (polynomial for natural frequency, and linear for damping), in Figures \ref{fig:wn_least_squares} and \ref{fig:zeta_least_squares}. In the figures, the ground-truth estimates are plotted as asterisks, and those associated with the FRFs used to train the combined-inference model are shown in red, while those at `unseen' temperatures are shown in black. The residue\textbackslash mode shape was assumed to be constant with respect to temperature.
	
	Figure \ref{fig:wn_least_squares} shows that a second-order polynomial appears to be an appropriate fit for natural frequency in the temperature range of interest. From Figure \ref{fig:zeta_least_squares}, a linear fit was considered most appropriate for damping. Although there may be a (weakly) nonlinear relationship between temperature and damping for this dataset, a linear assumption provided good results for the experiments used here. A higher-order Taylor series expansion for the temperature-damping relationship would require many coefficients, which would significantly increase the number of hyperparameters in the model. Further studies, including material-properties evaluation, may help elucidate the nature of this relationship specific to the helicopter blades, which may allow for better inferences from increased physics-based knowledge.
	
	\subsection{Model development and results} 
	\label{Case2_model}
	As with the first case, the population model for the second case included four latent functions (i.e., $ K = 4 $), as four temperature-varied FRFs provided training data, i.e., those at -10, -5, 10, and 25$^{\circ}$C. 	In addition, the models developed were trained using data with added noise, to avoid over-fitting small deviations/lightly participating modes within the band of interest. This was accomplished by adding zero-mean Gaussian noise, with variance equal to 10\% of the absolute maximum amplitude of the FRF obtained at -10$^{\circ}$C, to the 20 time-data blocks prior to computing FRFs. Then, the time data were subsampled such that all four FRFs used in training were data rich (In other words, each FRF used in training had enough training points to fully characterise the mode in this band. Subsampling in this case was performed not to demonstrate shrinkage towards the data-rich groups, but to reduce the dimensionality of the training data). FRFs were then computed and averaged, and truncated to the desired bandwidth. This procedure resulted in 122 training points from each of the four temperature-varied FRFs. Also, once again, only the real part of the FRF was fitted for simplicity. (Again, given that the imaginary part could be modelled in the same fashion, or inferred from the real part, focussing on the real part only was deemed sufficient for demonstrating the techniques presented herein.) The real components of the FRFs were modelled probabilistically with an assumed Student's t-distribution, using the same accelerance FRF estimation as a likelihood function, as in Eqs.\ \ref{eq:modalFRFreal} and \ref{eq:Hdist}. (Note that as with the previous case, $ \nu $ was assumed to be four. This assumption was made for consistency with the previous case, although it could be learnt as a model parameter if desired.) The additive noise variance, $ \sigma^2 $, of the FRFs was assumed shared among each of the domains (again, the same data acquisition system and sensors were used among the different tests, and the same Gaussian noise were added), and the modal residue was assumed constant with respect to temperature.
	
	Shared, higher-level distributions were placed over natural frequency, $ \boldsymbol{\omega}_{nat} = \{\omega_{nat}^{k}\}_{k=1}^4 $, damping, $ \boldsymbol{\zeta} = \{\zeta_{k}\}_{k=1}^4 $, and residue, $ A $, to allow information sharing among the domains,
	
	\begin{equation}
		\begin{split}
			&	\mu_{A} \sim \mathcal{N}\left(\text{-}0.008, 0.002^2\right), \sigma_{A}^2 \sim \mathcal{TN}\left(0.002, 0.002^2\right) \\ 
			&	\mu_{\omega_{nat}} \sim \mathcal{TN}\left(910, 10^2\right) \\ 	 
			&	\mu_{\zeta} \sim \mathcal{TN}\left(0.01, 1^2\right) \\ 	 
		\end{split}
		\label{eq:temps_popdist}
	\end{equation}
	
	\noindent Note that hyperpriors for $ \boldsymbol{\omega}_{nat} $ are shown in rad/s. Also note that for this preliminary model, normal distributions were assumed for all parameters (with truncated normal distributions assumed for variance, damping, and natural frequency). The only exception is the noise variance, which was modelled as a half-Cauchy, as with the previous case. The residue, $ A $, was assumed constant over all temperatures, with distributions,
	
	\begin{equation}
		A \sim \mathcal{N}\left(\mu_{A}, \sigma^2_{A} \right) \\
		\label{eq:temps_domainA}
	\end{equation}
	
	\noindent A second-order Taylor series expansion was used to approximate the functional relationship between temperature and natural frequency, with coefficients $ \mathbf{a} = \{a_1, a_2\} $, where the domain-level natural frequencies were defined as temperature-shifted realisations from the population-level distributions, via,
	\begin{equation}
		\{\omega_{nat}^k = \mu_{\omega_{nat}} + a_1T_k + a_2T_k^2\}_{k=1}^4 \\
		\label{eq:temps_domainfreq}
	\end{equation}

	\noindent Likewise, the relationship between temperature and modal damping was approximated as linear, with slope $ b $, over the measured temperatures, via,  
	
	\begin{equation}
		\{\zeta_{k} = \mu_{\zeta} + bT_k\}_{k=1}^4 \\
		\label{eq:temps_domainzeta}
	\end{equation}
	
	\noindent where $T_k$ was the temperature associated with the $k$th FRF. Priors were placed over the shared polynomial coefficients, with assumed distributions,
	
	\begin{equation}
		\begin{split}
			&	a_1 \sim \mathcal{N}\left(\text{-}0.01, 1^2\right) \\
			&	a_2 \sim \mathcal{N}\left(0.001, 1^2\right) \\
			&	b \sim \mathcal{N}\left(\text{-}5e{\text{-}6}, 1^2\right) \\
		\end{split}
		\label{eq:coeff_distributions}
	\end{equation}

	The shared noise variance, $ \sigma^2 $, was sampled from a half-Cauchy parent distribution, with higher-level scale parameter, $ \gamma^2 $, sampled from a truncated normal distribution,
	
	\begin{equation}
		\begin{split}
			&	\sigma^2 \sim \mathcal{C}\left(0,\gamma^2\right) \\
			&	\gamma^2 \sim \mathcal{TN}\left(0.01,0.05^2\right)\\ 	 
		\end{split}
		\label{eq:case2_noise_variance}
	\end{equation}
	
\begin{figure}[h]
	\centering
	\begin{tikzpicture}[latent/.style={circle, draw, minimum size=1.25cm}]
		\node[obs] (H) {$H_{i,k}$};
		\node[obs,right=1cm of H] (f) {$\omega_{i,k}$};
		\node[latent,above=1cm of H,xshift=2cm] (z) {$\zeta_{k}$}; 
		\node[latent,above=1cm of H,xshift=-2cm] (w) {$\omega_{nat}^k$}; 
		\node[latent,right=1cm of z] (A) {$A$}; 
		\node[latent,above=1cm of w,xshift=-2cm] (mu_w) {$\mu_{\omega_{nat}}$}; 
		\node[latent,above=1cm of w] (a) {$\mathbf{a}$}; 
		\node[obs,above=1cm of H] (T) {$T_k$}; 
		\node[latent,above=1cm of z,xshift=2cm] (mu_z) {$\mu_{\zeta}$}; 
		\node[latent,above=1cm of z] (b) {$b$};
		\node[latent,above=0.5cm of A,xshift=2cm] (mu_A) {$\mu_{A}$}; 
		\node[latent,below=0.5cm of mu_A] (sig_A) {$\sigma^2_{A}$};  
		\node[latent,left=3cm of H,yshift=0.6cm] (sig_H) {$\sigma^2$}; 
		\node[latent,above=1cm of sig_H] (gamma_sig) {$\gamma^2$}; 
		\node[const,below=0.6cm of sig_H] (nu) {$\text{ }\nu \text{ }$}; 
		\plate [inner sep=.25cm,yshift=.1cm,xshift=-.1cm] {plateN} {(H)(f)} {$i \in 1:N_k$};
		\plate [inner sep=.25cm,yshift=.1cm,xshift=0cm] {plateK} {(plateN)(w)(z)(T)} {$k \in 1:K$};
		\edge {w,A,z,f,sig_H} {H} 
		\edge {mu_w,T,a} {w} 
		\edge {mu_z,T,b} {z}
		\edge {mu_A,sig_A} {A}
		\edge {gamma_sig} {sig_H}
		\edge {nu} {H}
	\end{tikzpicture}
	\caption{Graphical representation of hierarchical FRF model for the second case.}
	\label{fig:DGM_temps}
	\end{figure}
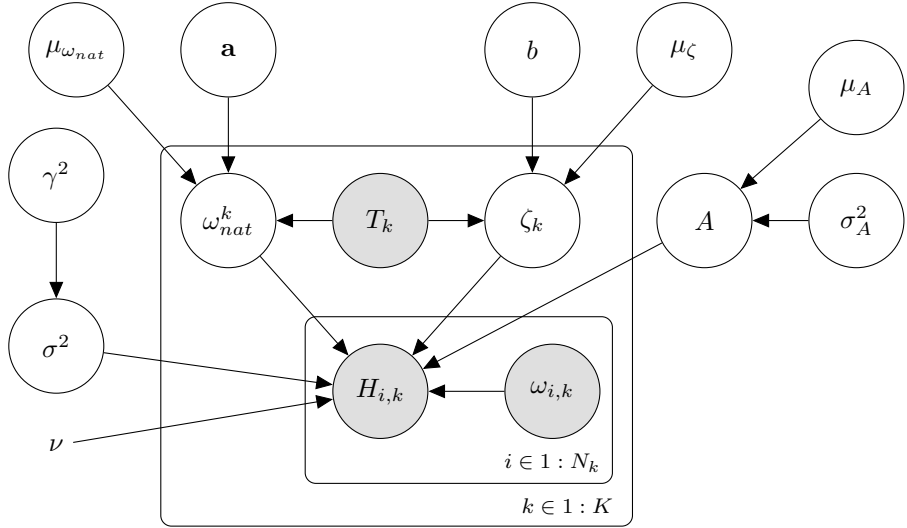

 	\noindent A graphical model displaying the parameter hierarchy is shown in Figure \ref{fig:DGM_temps}.
	
	As stated previously, FRFs captured at temperatures -10, -5, 10, and 25$^{\circ}$C provided the training data for the model. The intent was to use a sub-set of the temperature-varied FRFs, to learn the population-level coefficients necessary to describe the relationships between temperature and the modal parameters, so that these population-level variables can be used to make predictions at `unseen' temperatures (of course, these predictions were validated using experiments that did not contribute to the training data). As with the previous case, the \texttt{Stan} HMC sampler was run using four chains (and with a target average proposal acceptance probability rate of 0.8), for 10000 samples per chain (with an additional 5000 warm-up samples per chain, which were discarded), for each parameter. Posterior predictive checks were performed to ensure stationarity and proper mixing of the Markov chains. As computed by \texttt{Stan}, the \emph{maximum} split $ R_{hat} $ diagnostic, and the \emph{minimum} effective sample size ($ N_{Eff} $), considering all parameters/hyperparameters, are shown in Table \ref{tab:MCMCchecks_Case2}, for all models (including those used for `ground truth' natural frequency and damping estimates). Note that split $ R_{hat} $ values did not exceed 1.01, and $ N_{Eff} $ was greater than 400, suggesting proper mixing of the Markov chains, per current guidelines \cite{Vehtari2021}.
	
	\begin{table}[h]
		\renewcommand{\arraystretch}{1.5}
		\caption{\label{tab:MCMCchecks_Case2}Posterior predictive diagnostics for Case 2.}
		\begin{center}
			\begin{tabular}{ p{1.6cm} p{2.5cm} p{2.5cm} }
				\hline
				Metric & Partial-Pooling Model & `Ground-Truth' Models \\
				\hline
				min $ N_{Eff} $ & 15000 & 640 \\
				max $ R_{hat} $  & 1 & 1 \\
				\hline
			\end{tabular}
		\end{center}
	\end{table}

\begin{figure}
		\centering
		{\includegraphics[width=0.8\textwidth]{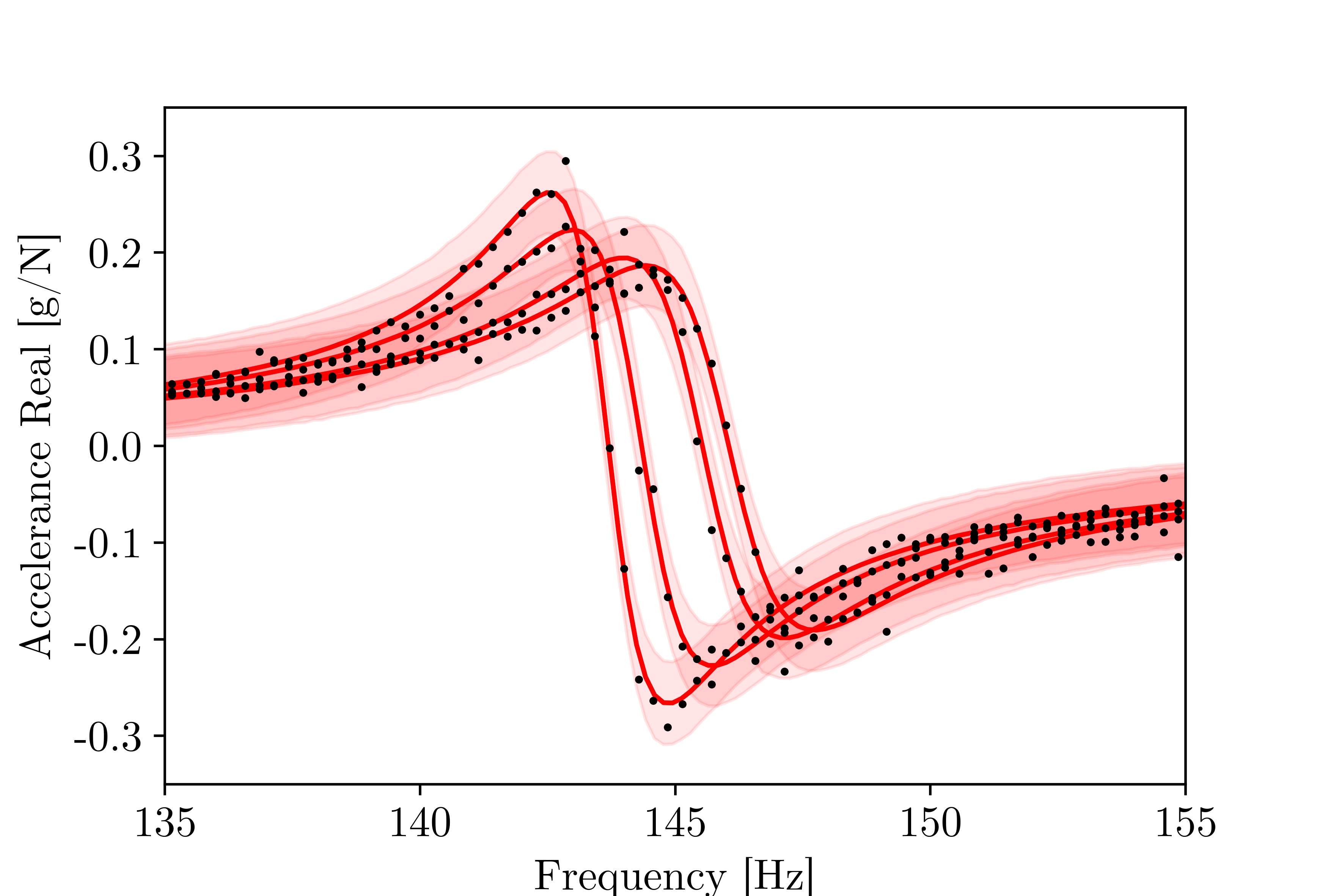}}
		\captionof{figure}{Posterior predictive mean and 3-sigma deviation for probabilistic FRF model, for training data at -10, -5, 10, and 25$^{\circ}$C. Training data are shown in a black scatter plot. Solid lines are the posterior predictive means, and the variance is represented by shaded regions.}
		\label{fig:temps_modelFRFs}
	\end{figure}

	Eq.\ (\ref{eq:modalFRFreal}) was used to compute FRFs from the posterior MCMC samples of the modal parameters, for each temperature-varied FRF used in training. FRFs \emph{with predicted variance} were generated by sampling from a Student's t-distribution with four degrees-of-freedom, mean equal to the computed FRFs, and posterior predictive measurement noise, $ \sigma^2 $, for each MCMC sample. Total variance was estimated by taking the standard deviation of the FRFs. Posterior predictive means and 3-sigma deviation for the FRFs are shown in Figure \ref{fig:temps_modelFRFs}.
	
	The expectations for the population-level parameters were obtained from the samples, including the polynomial coefficients that define the relationships between temperature and the modal parameters. The expectations are shown in Table \ref{tab:population_means}. The procedure for extrapolating beyond the FRFs used in training is described in Section \ref{Case2_extrapolation}.
	
	\begin{table}[h]
		\renewcommand{\arraystretch}{1.5}
		\caption{\label{tab:population_means}Expectation of the population-level variables.}
		\begin{center}
			\begin{tabular}{ p{1.2cm} p{1.4cm} p{1.1cm} p{1.2cm} p{1.1cm} p{1.4cm} }
				\hline
				$\text{E}[A]$ & $\text{E}[{\mu}_{\omega_{nat}}]$ &  $\text{E}[{\mu}_{\zeta}]$ & $\text{E}[{a_1}]$ & $\text{E}[{a_2}]$ & $\text{E}[{b}]$ \\
				\hline
				-0.0083&145.1&0.0104&-0.5360&0.0075&-9.272e{-5}\\
				\hline
			\end{tabular}
		\end{center}
	\end{table}

	\subsection{Extrapolation to other temperatures} \label{Case2_extrapolation}
	
	After training the model, predictions were made at `unseen' temperatures, i.e., those not used in training. These predictions were made by computing natural frequency and damping using posterior MCMC samples, for all the measured temperatures, $ T = \{T_i\}_{i=1}^{11} $, including the four used in training, via the relations,
	
	\begin{equation}
		\begin{split}
			&	\{\{\omega_{nat}^{i,n} = {\mu}_{\omega_{nat,n}} + {a_{1,n}}T_i + {a_{2,n}}T_i^2\}_{i=1}^{11}\}_{n=1}^{40000} \\
			&	\{\{\zeta_{i,n} = {\mu}_{\zeta,n} + {b_n}T_i\}_{i=1}^{11}\}_{n=1}^{40000} \\
		\end{split}
		\label{eq:extrapolation_params}
	\end{equation}
	
\begin{figure}
		\centering
		{\includegraphics[width=0.8\textwidth]{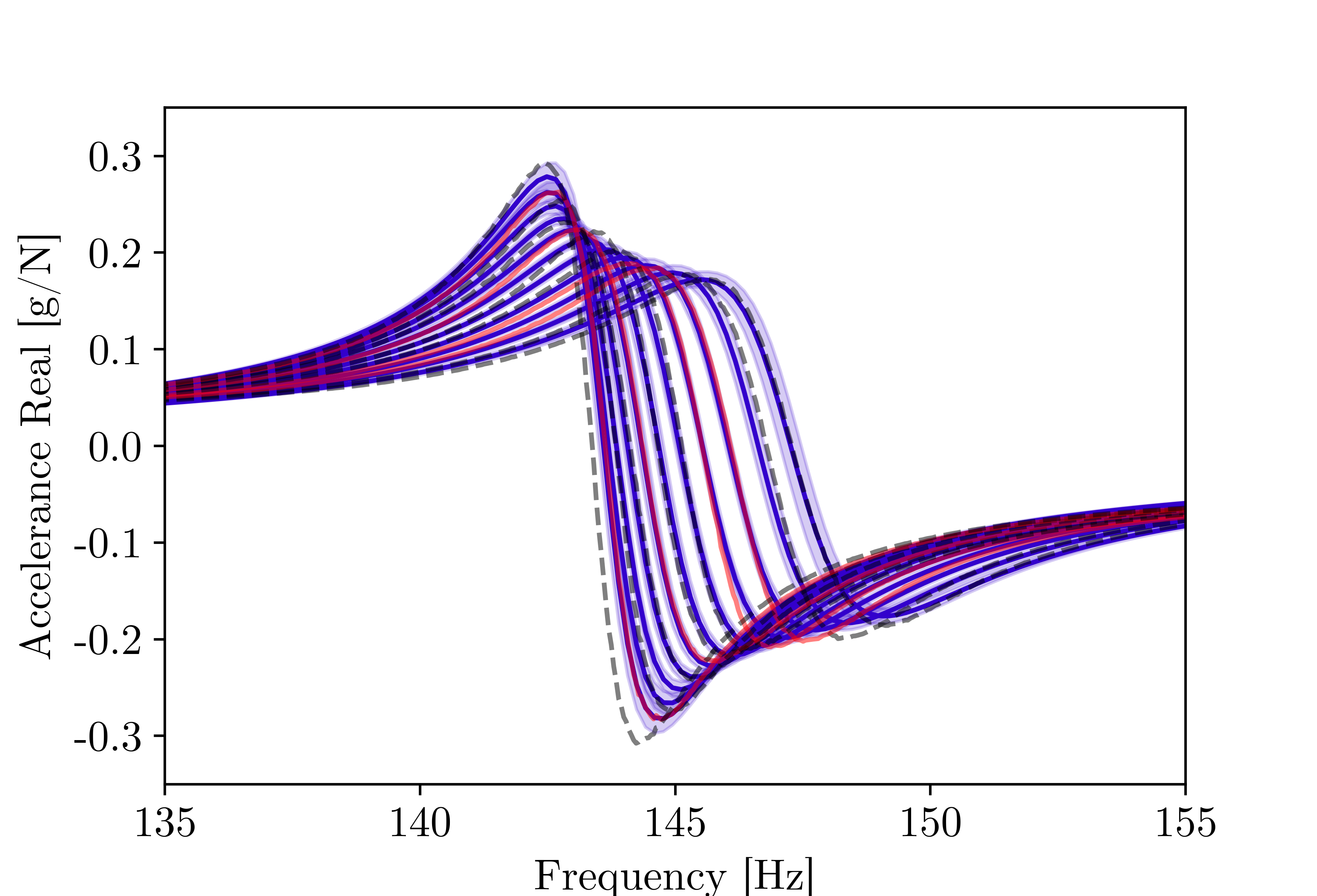}}
		\captionof{figure}{Extrapolated FRFs using population-level variables, from -20 to 30$^{\circ}$C in increments of 5$^{\circ}$C, with decreasing temperature from left to right. The extrapolated FRF means are shown as solid blue lines, with shaded blue regions indicating variance bounds (3-sigma deviation). Solid red lines denote the FRFs that provided training data for the model, while dashed black lines denote the measured FRFs at `unseen' temperatures.}
		\label{fig:extrapolatedFRFs}
	\end{figure}

\begin{figure}[h!]
		\centering
		\subfloat[\label{fig:extrapolated_wn}]{\includegraphics[width=0.8\textwidth]{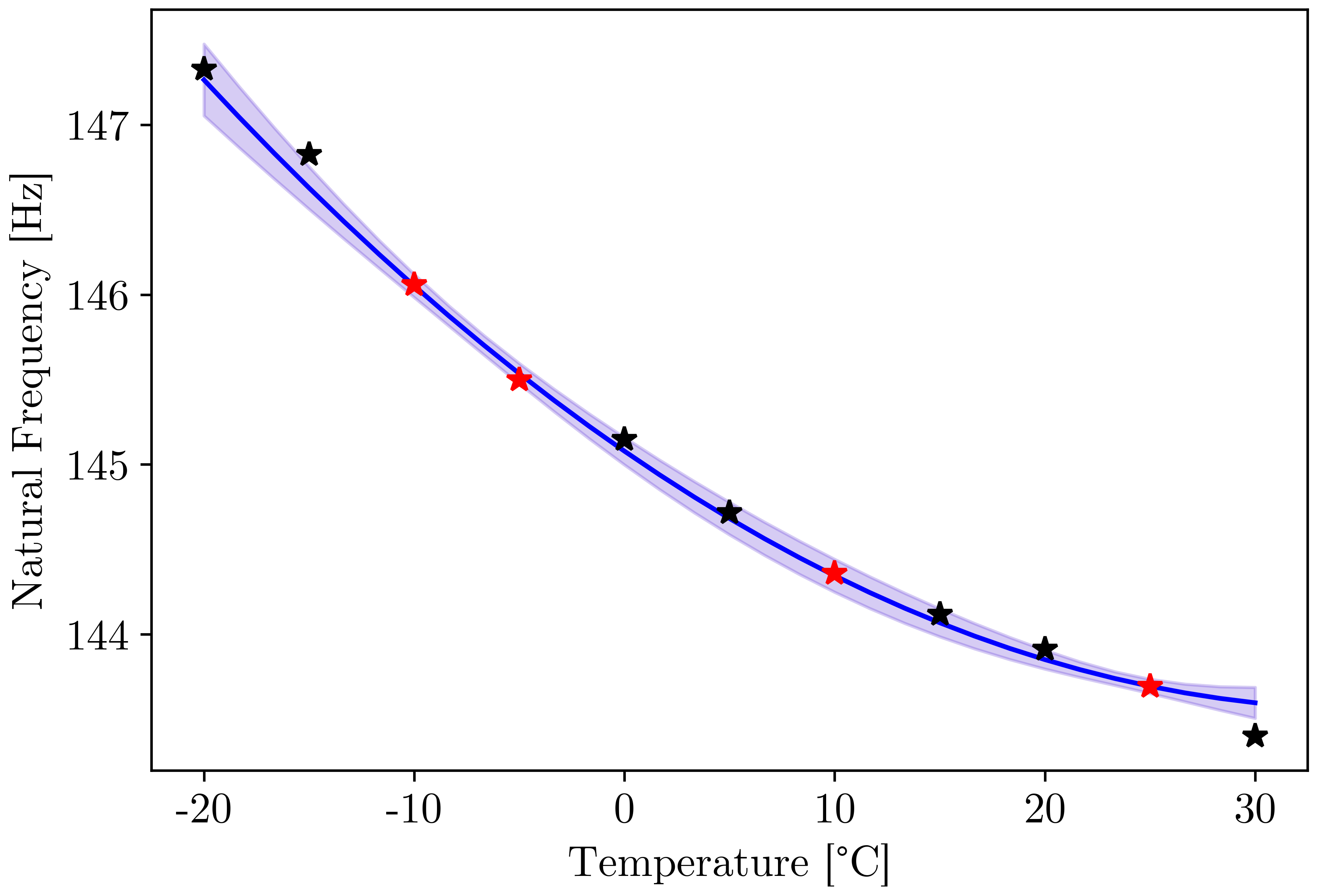}} \\
		\subfloat[\label{fig:extrapolated_zeta}]{\includegraphics[width=0.8\textwidth]{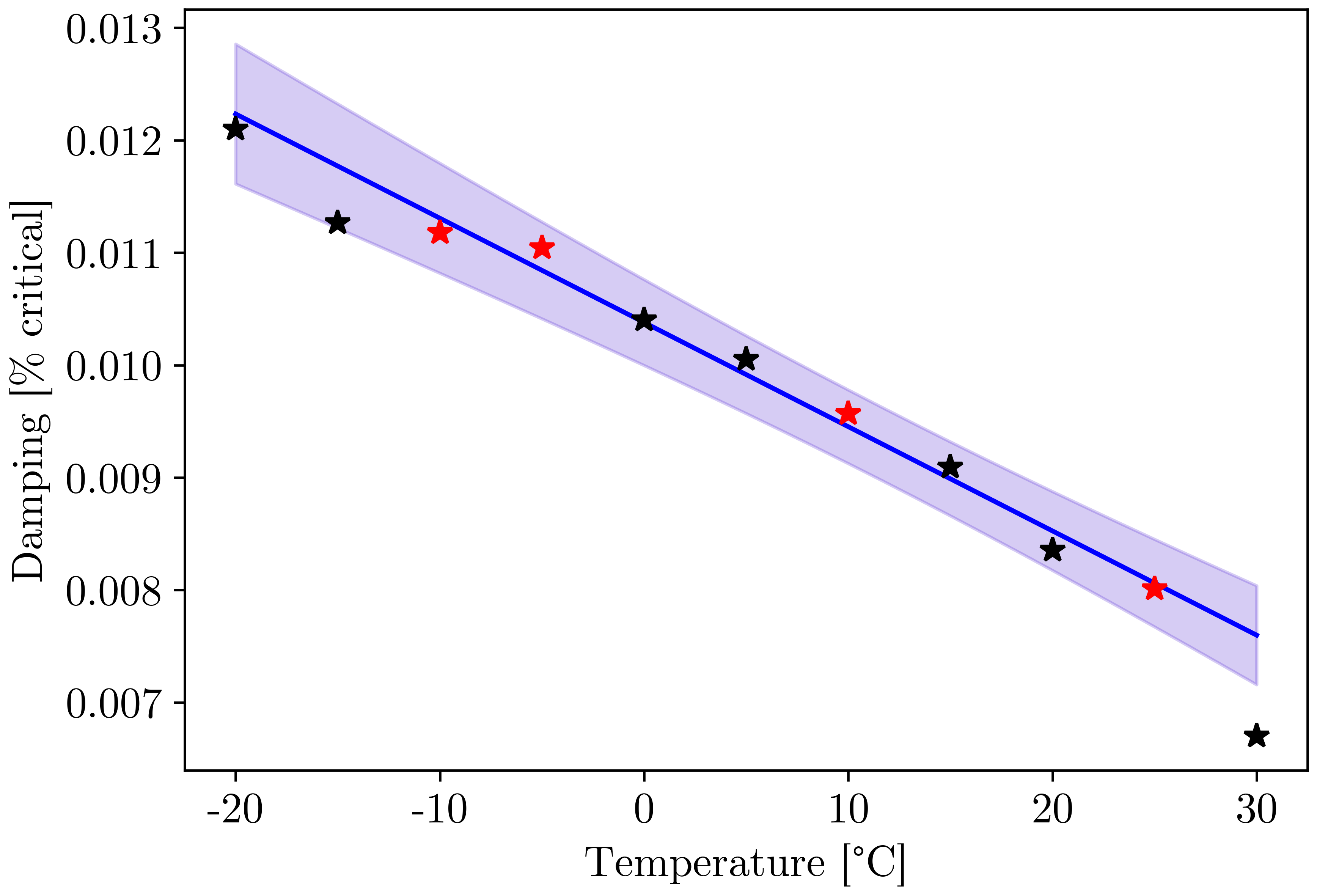}} \\
		\caption{Mean extrapolated (a) natural frequency and (b) damping for Blade 1, at training and `unseen' temperatures, plotted against temperature, with 3-sigma deviation (shown in blue). Ground-truth model variables are shown as stars, with red indicating the temperatures used in training and black denoting `unseen' temperatures.}
	\end{figure}

	\noindent Natural frequency and damping were computed for each each MCMC sample, for each temperature. These values were then used to compute FRFs via Eq.\ (\ref{eq:modalFRFreal}). Variance was estimated by taking the standard deviation of the FRFs. The extrapolated FRF means are plotted in Figure \ref{fig:extrapolatedFRFs}, along with 3-sigma deviation. The extrapolated FRFs shown in Figure \ref{fig:extrapolatedFRFs} range from -20 to 30$^{\circ}$C in increments of 5$^{\circ}$C, with decreasing temperature from left to right, as natural frequency increases with decreasing temperature. Model accuracy was evaluated by calculating the NMSE via Eq.\ (\ref{eq:NMSE}) for each measured FRF without added noise, compared to the mean extrapolated FRFs, as shown in Table \ref{tab:NMSE}. Note that the NMSE computed for the prediction at temperatures -10, -5, 10, and 25$^{\circ}$C, which correspond to the data used in model training, are included in the table for completeness. However, the specific training points were excluded from the calculation. To further visualise the relationships between temperature and the modal parameters, the functions were plotted for each sample. The mean and variance of the functions were plotted with the parameters estimated via independent modelling (in this case, these results were considered to be the best approximation of the ground truth, as each FRF was data-rich). These plots are shown in Figures \ref{fig:extrapolated_wn} and \ref{fig:extrapolated_zeta}. 
		
	\begin{table}[h]
		\renewcommand{\arraystretch}{1.5}
		\caption{\label{tab:NMSE}Evaluation of model accuracy for extrapolation to `unseen' data.}
		\begin{center}
			\begin{tabular}{ p{3cm} p{1.5cm}}
				\hline
				Temperature [$^{\circ}$C] & NMSE \\
				\hline
				-20 & 0.36 \\
				-15 & 0.86 \\
				-10 & 0.46 \\
				-5 & 0.59 \\
				0 & 0.68 \\
				5 & 0.52 \\
				10 & 0.42 \\
				15 & 0.47 \\
				20 & 0.67 \\
				25 & 0.80 \\
				30 & 3.36 \\
				\hline
			\end{tabular}
		\end{center}
	\end{table}
	
	In Figure \ref{fig:extrapolatedFRFs}, the extrapolated FRFs are shown as solid blue lines, with a shaded blue region indicating the variance bounds. The measured FRFs used to train the model, without added noise, are shown as solid red lines, and the measured FRFs used to test the model at `unseen' temperatures are shown as dashed black lines. From the figure, it is clear that excellent agreement was achieved between the extrapolated and measured FRFs at the temperatures used to train the model, as expected. Likewise, the FRFs at temperatures between those used to train the model (e.g., at 0, 15, and 20$^{\circ}$C) show excellent agreement. Notably, the FRFs at colder temperatures, which were further away from the training data, still show good agreement. The accuracy of the fit is further shown in Table \ref{tab:NMSE}, as all NMSE values were less than 5\%.
	
	Figures \ref{fig:extrapolated_wn} and \ref{fig:extrapolated_zeta} further show that the extrapolated parameters accurately represented the measured data, as the measured parameters largely fell within the variance bounds of the approximations. Figure \ref{fig:extrapolated_wn} shows that for natural frequency, parameter variance increased as the inferred parameters varied further from the training data (i.e., at temperatures further from those used to train the model), as expected. Figure \ref{fig:extrapolated_zeta} shows that for damping, the relatively consistent variance was the result of the linear assumption for the temperature-damping relationship, and the relatively high deviation of the measured data from the linear fit. Further model development could involve the incorporation of more physics-based knowledge, such as, forcing the relationship between natural frequency and temperature to be monotonically decreasing, or further investigation into the possibly nonlinear relationship between damping and temperature. In addition, it may be possible to augment the model by imposing a sparsity-inducing prior on the weights of the function describing the relationship. This sparsity-inducing prior would allow the Bayesian inference to perform system identification with automatic relevance detection, allowing incorporation of additional terms, polynomial or otherwise.

\vspace{12pt} 
\section{Concluding remarks} \label{conclusions}
	
	Current work has involved the development of probabilistic FRF models using a hierarchical Bayesian approach, that account for benign variations as well as similarities among nominally-identical structures. Two cases were presented, to demonstrate the usefulness of this modelling structure. The first case demonstrated how hierarchical Bayesian models with partial pooling can reduce variance in data-poor domains by allowing information transfer with data-rich domains, via shared population-level distributions. The second case showed that incorporating functional relationships (approximated via Taylor series expansion) into the modelling structure to describe temperature variations allows for prediction beyond the training data. The first case addresses data-sparsity challenges in SHM, as missing time-domain data resulting from sensor dropout and other causes will reduce the number of spectral lines in the frequency domain, which can impede dynamic characterisation. Likewise, the second case addresses data-scarcity issues, as encoding physics-based knowledge into the model via functional relationships can increase the amount of normal-condition information available. Future efforts may involve further investigation into the physical relationships between temperature and the modal parameters, particularly damping, to improve model accuracy. 
	
\vspace{12pt} 
\section{Acknowledgements} \label{acknowledgements}
	The authors gratefully acknowledge the support of the UK Engineering and Physical Sciences Research Council (EPSRC), via grant reference EP/W005816/\-1. This research made use of The Laboratory for Verification and Validation (LVV), which was funded by the EPSRC (via EP/J013714/1 and EP/N010884/1), the European Regional Development Fund (ERDF), and the University of Sheffield. For the purpose of open access, the authors have applied a Creative Commons Attribution (CC BY) licence to any Author Accepted Manuscript version arising. The authors would like to extend special thanks to Michael Dutchman of the LVV, for helping set up the experiments, and also Domenic Di Francesco of the Alan Turing Institute, for his advice when designing the hierarchical models.


	
\appendix
\counterwithin{figure}{section}
\section{One data-rich FRF, three data-poor FRFs}
\label{appendix}
\begin{figure}[H]
		\centering
		{\includegraphics[width=0.8\textwidth]{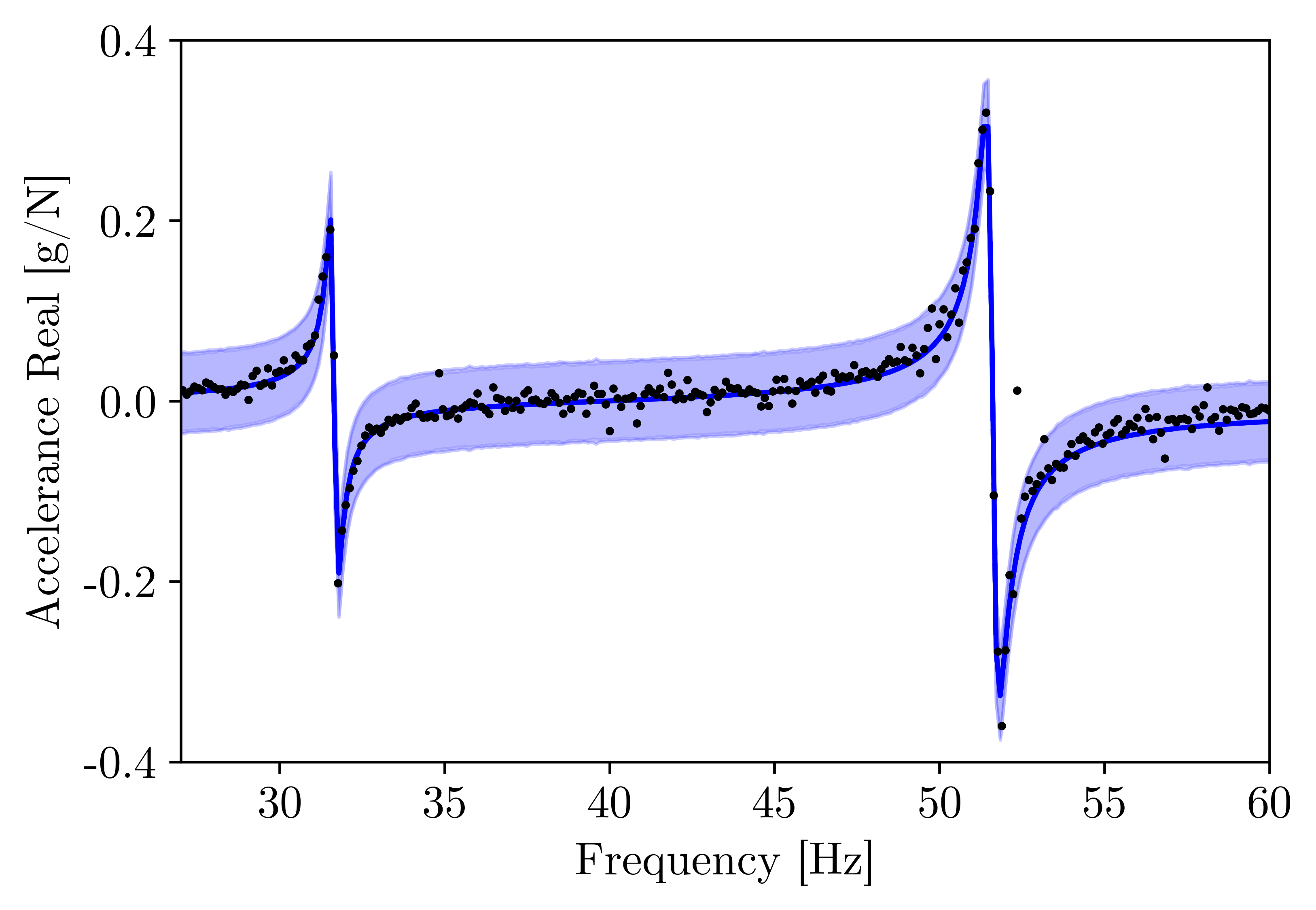}} \\
		\caption{Posterior predictive mean and 3-sigma deviation for independent (no-pooling) and partial-pooling models, for data-rich Blade 1. Training data are shown in a black scatter plot. The posterior predictive means for the partial-pooling and independent models are shown as solid and dashed lines, respectively.  The variance is represented by shaded regions, where the independent model variance is shown in a lighter colour than that for the partial-pooling model.}
		\label{fig:B1_appendix}
	\end{figure}

\begin{figure}
		\centering
		{\includegraphics[width=0.8\textwidth]{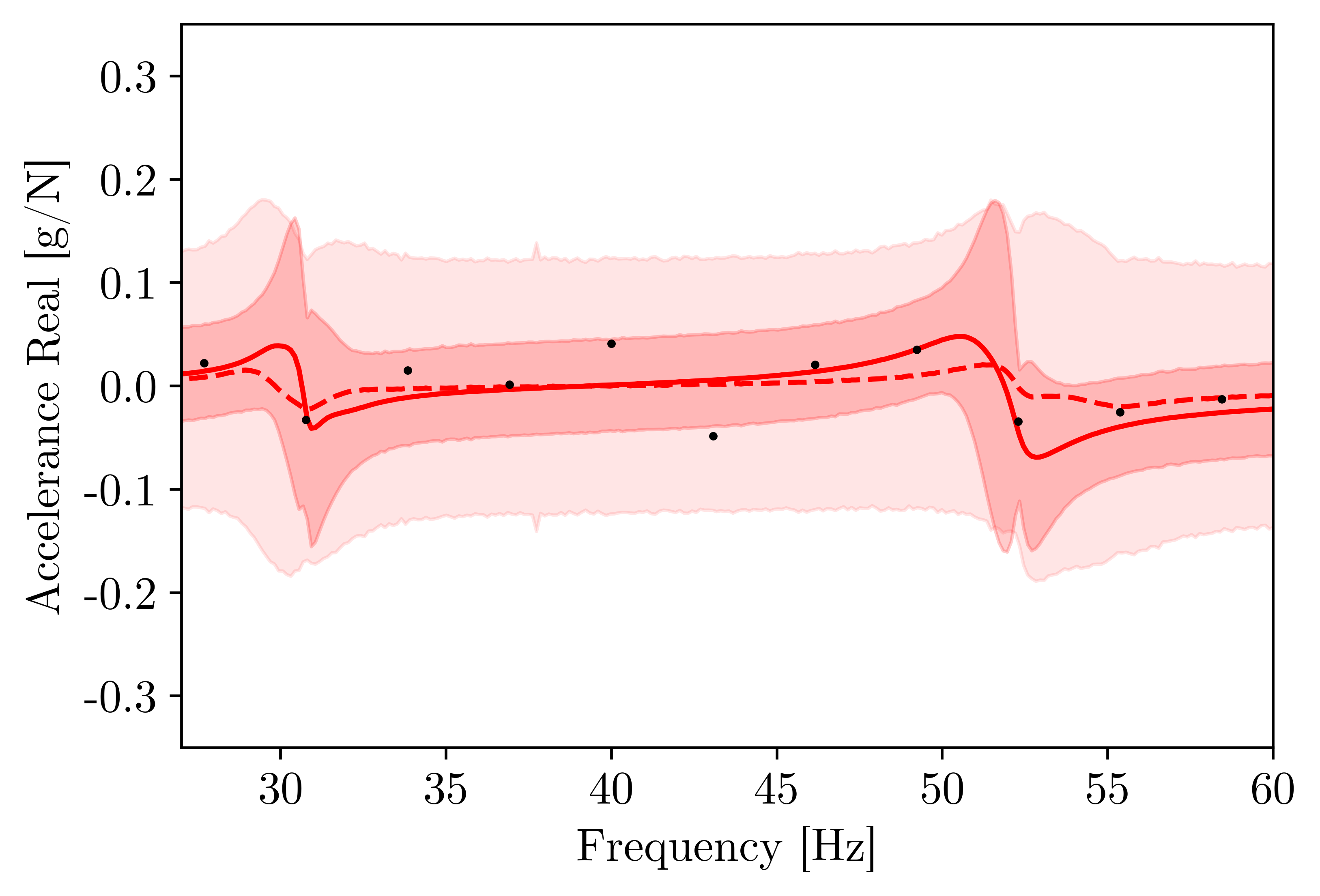}} \\
		\caption{Posterior predictive mean and 3-sigma deviation for independent (no-pooling) and partial-pooling models, for data-poor Blade 2. Training data are shown in a black scatter plot. The posterior predictive means for the partial-pooling and independent models are shown as solid and dashed lines, respectively.  The variance is represented by shaded regions, where the independent model variance is shown in a lighter colour than that for the partial-pooling model.}
		\label{fig:B2_appendix}
	\end{figure}

\begin{figure}
		\centering
		{\includegraphics[width=0.8\textwidth]{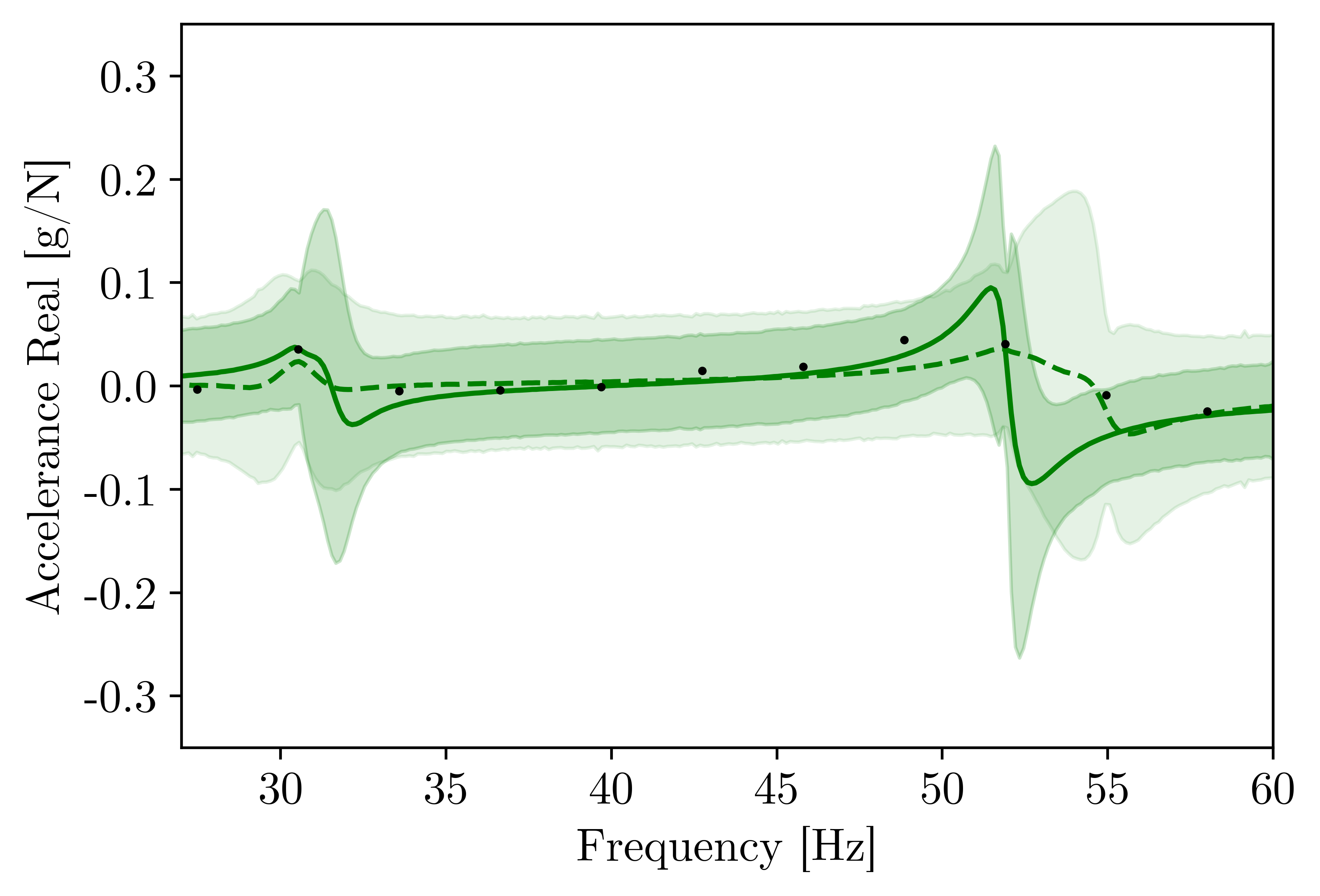}} \\
		\caption{Posterior predictive mean and 3-sigma deviation for independent (no-pooling) and partial-pooling models, for data-poor Blade 3. Training data are shown in a black scatter plot. The posterior predictive means for the partial-pooling and independent models are shown as solid and dashed lines, respectively.  The variance is represented by shaded regions, where the independent model variance is shown in a lighter colour than that for the partial-pooling model.}
		\label{fig:B3_appendix}
	\end{figure}

\begin{figure}[H]
		\centering
		{\includegraphics[width=0.8\textwidth]{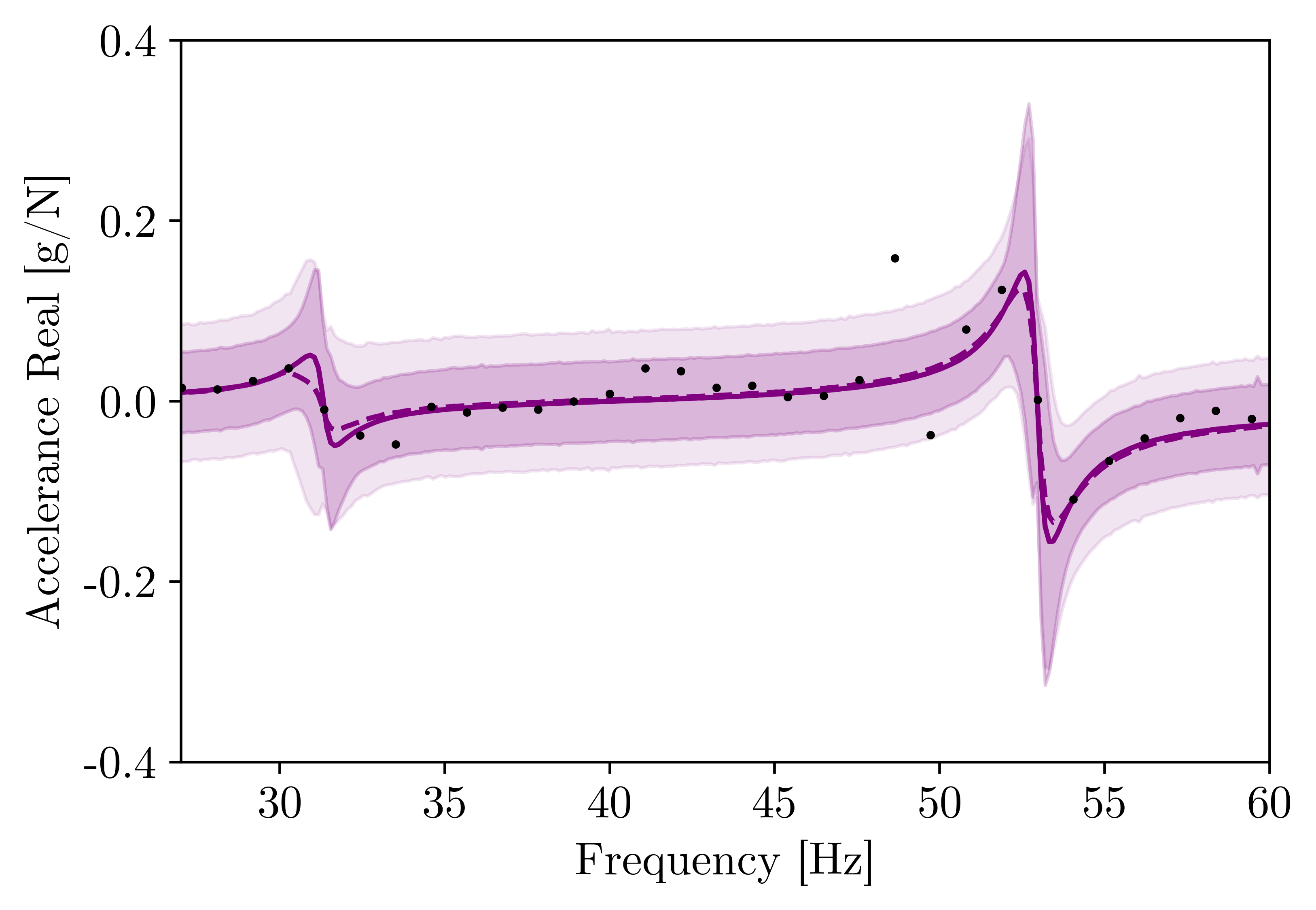}} \\
		\caption{Posterior predictive mean and 3-sigma deviation for independent (no-pooling) and partial-pooling models, for data-poor Blade 4. Training data are shown in a black scatter plot. The posterior predictive means for the partial-pooling and independent models are shown as solid and dashed lines, respectively.  The variance is represented by shaded regions, where the independent model variance is shown in a lighter colour than that for the partial-pooling model.}
		\label{fig:B4_appendix}
	\end{figure}

	Consider a situation, where comprehensive data were only known for a single structure in the population, with sparse data available for the remaining three structures. As with the case where two structures were data rich, and two were data poor, the result was shrinkage towards the population mean. The population mean was largely informed by the data-rich group, and prior information, with varying contributions from the sparse groups depending on how informative the available data were. Posterior predictive mean and 3-sigma deviation for the partial-pooling and independent models are plotted in Figures \ref{fig:B1_appendix} to \ref{fig:B4_appendix}, respectively. Note that the variance reduction was especially significant for Blade 2, as the data for this blade were quite uninformative.
	
\end{document}